\documentclass[10pt,twocolumn,letterpaper]{article}

\usepackage{cvpr}              % To produce the CAMERA-READY version
\usepackage{multirow}
\usepackage[normalem]{ulem}

\usepackage[accsupp]{axessibility}

\usepackage{algorithm}
\usepackage{algpseudocode}
\usepackage{booktabs}              % 提供 \toprule \midrule \bottomrule
\usepackage[table]{xcolor}         % 提供 \rowcolor（内部加载 colortbl）
\usepackage{tabularx}
\definecolor{cvprblue}{rgb}{0.21,0.49,0.74}
\usepackage[pagebackref,breaklinks,colorlinks,allcolors=cvprblue]{hyperref}

\def\paperID{} % *** Enter the Paper ID here
\def\confName{CVPR}
\def\confYear{2026}

\title{Towards Stable Self-Supervised Object Representations in Unconstrained Egocentric Video}

\author{Yuting Tan$^{1,*}$, Xilong Cheng$^{1,*}$, Yunxiao Qin$^{1,2,*,\dagger}$, Zhengnan Li$^{3}$, Jingjing Zhang$^{1,\dagger}$\\
$^{1}$Communication University of China, Beijing, China\\
$^{2}$State Key Laboratory of Media Convergence and Communication, Beijing, China\\
$^{3}$The Chinese University of Hong Kong, Shenzhen\\
{\tt\small \{yutingtan, chengzhengyu330, qinyunxiao\}@cuc.edu.cn}
}

\begin{document}
\maketitle

\renewcommand{\thefootnote}{\fnsymbol{footnote}} % 切换为符号模式
\footnotetext[1]{Equal contribution} % 对应 *
\footnotetext[2]{Corresponding authors} % 对应 dagger

\begin{abstract}
Humans develop visual intelligence through perceiving and interacting with their environment—a self-supervised learning process grounded in egocentric experience. 
Inspired by this, we ask how can artificial systems learn stable object representations from continuous, uncurated first-person videos without relying on manual annotations.
This setting poses challenges of separating, recognizing, and persistently tracking objects amid clutter, occlusion, and ego-motion.
We propose EgoViT, a unified vision Transformer framework designed to learn stable object representations from unlabeled egocentric video. 
EgoViT bootstraps this learning process by jointly discovering and stabilizing "proto-objects" through three synergistic mechanisms: (1) Proto-object Learning, which uses intra-frame distillation to form discriminative representations; (2) Depth Regularization, which grounds these representations in geometric structure; and (3) Teacher-Filtered Temporal Consistency, which enforces identity over time. 
This creates a virtuous cycle where initial object hypotheses are progressively refined into stable, persistent representations. 
The framework is trained end-to-end on unlabeled first-person videos and exhibits robustness to geometric priors of varied origin and quality.
On standard benchmarks, EgoViT achieves +8.0\% CorLoc improvement in unsupervised object discovery and +4.8\% mIoU improvement in semantic segmentation, demonstrating its potential to lay a foundation for robust visual abstraction in embodied intelligence.
\end{abstract}    
\section{Introduction}
\label{sec:intro}
\begin{figure}
    % \vspace{-24pt}
    \centering
    \includegraphics[width=\linewidth]{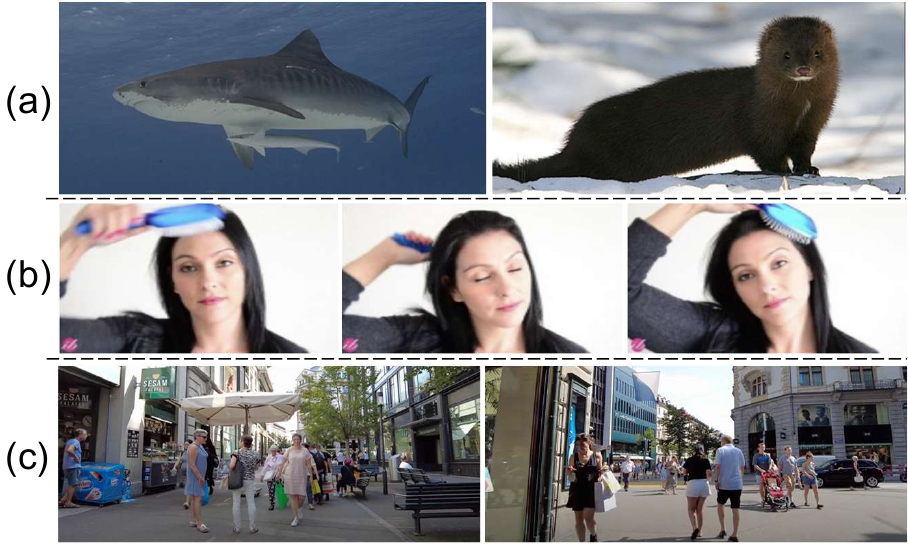}
    \vspace{-20pt}
    \caption{ 
    Visual data complexity comparison. 
    (a) ImageNet  \cite{deng2009imagenet}and (b) Kinetics-400 \cite{kay2017kinetics} feature predominantly object-centric scenes with clear backgrounds or structured interactions. 
    (c) Unconstrained egocentric videos \cite{venkataramanan2023imagenet} present substantially greater complexity, featuring dense object interactions, severe occlusions, and continuous ego-motion, which together pose unique challenges for learning persistent object representations.
    }
    \label{fig:object_centric_comparison}
    \vspace{-8pt}
\end{figure}

Human visual intelligence is forged through egocentric embodied experience, a powerful form of self-supervised learning \cite{smith2005development}.
This process provides a continuous, temporally coherent visual stream that is rich with a wealth of supervisory signals about the physical world. 
By observing how objects persist, move, and interact over time, we learn robust vision intelligence and fundamental concepts like object permanence \cite{piaget2013construction} and spatiotemporal dynamics \cite{kellman1983perception}. 
However, the dominant paradigms in computer vision, while powerful, were not fundamentally designed to exploit such rich, continuous data.
They are mainly developed on datasets of static, third-person images with curated, center-biased object compositions \cite{deng2009imagenet, chen2020simple, he2020momentum, he2022masked} or short video clips featuring scripted actions in controlled environments \cite{kay2017kinetics, tong2022videomae}, as illustrated in Fig.\ref{fig:object_centric_comparison}(a) and (b).
Consequently, the resulting methods, while successful in many visual tasks, are not explicitly optimized to learn the long-horizon temporal coherence.
As a result, they often struggle to maintain stable object identities amidst the severe occlusions, continuous ego-motion, and cluttered scenes characteristic of unconstrained egocentric video \cite{grauman2022ego4d}.

Indeed, the very temporal richness that makes this data promising introduces unique challenges: constant viewpoint shifts and severe occlusions make it difficult to maintain consistent object representations (Fig.~\ref{fig:object_centric_comparison}(c)), a challenge shared with but distinct from traditional multi-object tracking due to the lack of pre-defined object detectors \cite{bewley2016simple, wojke2017simple}. 
This confluence of opportunity and difficulty defines our central research question: \emph{How can we learn identity-consistent object representations from complex, unconstrained egocentric videos without manual annotations?}

To address this, we propose \textbf{Ego}centric \textbf{Vi}sion \textbf{T}ransformer (EgoViT), a unified framework that shifts the learning focus from low-level pixel correspondence to the discovery and tracking of emergent proto-objects (emergent, reusable visual components) through its attention mechanisms.
EgoViT bootstraps this process through a virtuous cycle of three synergistic mechanisms:
(1) Proto-object Learning: Employs intra-frame distillation to learn discriminative representations;
(2) Depth Regularization: Grounds the representations in geometric reality by leveraging depth priors as structural constraints;
and (3) Teacher-Filtered Temporal Consistency: Leverages a stable teacher to prune unreliable correspondences caused by occlusion and ego-motion and enforce identity.

Extensive experiments on established benchmarks show that EgoViT substantially outperforms recent self-supervised baselines, achieving a gain of +8.0\% CorLoc in unsupervised object discovery and +4.8\% mIoU in semantic segmentation.
These results demonstrate the potent effect of synergistically integrating appearance, depth, and temporal cues for learning from egocentric video.
More broadly, our work represents a conceptual shift from static recognition of \textbf{what} objects are toward a dynamic understanding of \textbf{how} they persist over time.
This may lay a foundation for future embodied AI systems to build persistent world models.

In summary, our contributions are threefold.
\textbf{First}, we propose a novel synergistic stabilization approach for learning in unconstrained egocentric video.
We hypothesize that stable object representations can emerge from joint optimization of complementary appearance, depth, and temporal mechanisms.
\textbf{Second}, we present EgoViT, a unified teacher-student framework that instantiates this approach and demonstrates how these three mechanisms can be effectively integrated into a single, end-to-end trainable architecture.
\textbf{Finally}, we provide rigorous empirical validation for our formulation through extensive experiments and ablation studies.
The clear performance gains over strong baselines serve as compelling evidence for the efficacy and superiority of our proposed synergistic stabilization approach.

\vspace{-6pt}
\section{Related Work}
Our work bridges self-supervised video representation learning and unsupervised object discovery, tailored to the unique challenges of unconstrained egocentric vision.

\subsection{Self-Supervised \& Object-Centric Learning}
% \subsection{Self-Supervised Video Learning}
A primary strategy in self-supervised video learning is to leverage temporal coherence as a supervisory signal. 
Foundational approaches learn robust visual features by tracking image patches \cite{wang2015unsupervised}, verifying the temporal order of shuffled frames \cite{misra2016shuffle,lee2017unsupervised}, or using time-contrastive learning \cite{sermanet2018time, gordon2020watching,wang2024poodle}.
Other works exploit cross-modal signals like audio \cite{arandjelovic2017look,owens2018audio,wang2020self}, predict future frames \cite{srivastava2015unsupervised,mathieu2015deep,finn2016unsupervised}, or enforce equivariance to transformations like optical flow \cite{xiong2021self}.
By design, they learn to associate generic patches or frames, making them ill-suited for discovering the identity of discrete objects, a core challenge in cluttered egocentric streams.
To tackle this challenge directly, the paradigm of unsupervised object-centric learning has emerged.
However, its main strategies also face hurdles in egocentric video.
Iterative refinement methods, such as Slot Attention and its successors \cite{locatello2020object,kipf2021conditional,wu2022slotformer,elsayed2022savi++,van2024moving} assume a quasi-static input violated by first-person video's non-stationary nature \cite{kanade2012first, grauman2022ego4d}.
Concurrently, motion-based grouping\cite{chen2021roots} struggles to disentangle true object movement from the camera's significant ego-motion \cite{nagarajan2020ego, zadaianchuk2023object}. 
Other approaches introduce spatial locality priors~\cite{chakravarthy2023spotlight} or extend decomposition to 3D~\cite{henderson2020unsupervised}, but often operate in scenes that are less dynamic than the unconstrained videos we target. 
Our task also differs from traditional Multi-Object Tracking (MOT) \cite{bewley2016simple, leal2016learning, wojke2017simple,park2021multiple}, which relies on pre-defined, category-specific detectors and cannot discover novel objects.

While both EgoViT and the recent work DoRA~\cite{venkataramanan2023imagenet} leverage ViT attention for prototype discovery, they fundamentally diverge in their use of temporal information—the very critical dimension in egocentric video. 
DoRA employs temporal correspondence primarily for spatial data augmentation, creating masked views for its spatial consistency loss. 
In contrast, EgoViT introduces a direct proto-to-proto temporal alignment objective $\mathcal{L}_{\text{temp}}$ where time itself becomes the primary axis of supervision, a process uniquely stabilized by our depth regularization prior (see Appendix~\ref{appendix:Core-differences-between-EgoViT-and-DoRA} for a detailed comparison).

\vspace{-5pt}
\subsection{Biological Inspiration}
\vspace{-2pt}
Neuroscience research shows that stereoscopic vision and depth processing enhance the primate brain's ability to track dynamic targets \cite{gonzalez1998neural,wolbers2008spatial,allman1999evolving,julesz1971foundations}. 
Binocular disparities processed in the primary visual cortex (V1) and middle temporal area (MT) enable target-background discrimination \cite{roe2007disparity,deangelis1999organization}, with stereopsis loss severely impairing tracking \cite{zihl1983selective}.
Importantly, tracking refines object representations \cite{hochstein2002view,baillargeon1987object,kellman1983perception}, a process driven by dorsal-ventral stream interactions that build view-invariant codes \cite{booth1998view,logothetis1995shape} and guide attentional feedback \cite{hopfinger2000neural,crawford2011three}.

While our work draws from these biological principles, it is important to situate it relative to research inspired by developmental psychology and cognitive science. 
Pioneering studies have made remarkable progress by learning from a single child's first-person perspective \cite{orhan2020self, vong2024grounded} or by leveraging explicit interaction cues, such as object manipulation and gaze, as potent self-supervisory signals \cite{aubret2024self, aubret2022time, aubret2024learning}. 
A common thread in these important prior works is their reliance on either more structured visual data, such as in domestic environments with fewer salient objects, or explicit behavioral priors. 
Our work is designed precisely for the more challenging scenario of discovering and persistently tracking multiple objects amidst the dense, ``in-the-wild" clutter of unconstrained egocentric videos, relying instead on general visual cues like depth and temporal coherence.

\begin{figure}[t]
  \centering
  % \vspace{-18pt}
  \includegraphics[width=0.8\linewidth]{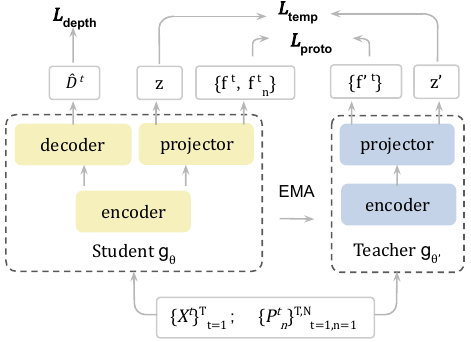}
  \vspace{-10pt}
  \caption{
    {EgoViT} adopts a Teacher-Student architecture, processing input frames $\{X^{t}\}_{t=1}^{T}$ and $\{P_n^t\}^{T,N}_{t=1,n=1}$.
    Student $g_\theta$ learns from three mechanism: (1) depth-regularization $\mathcal{L}_{\text{depth}}$;
    (2) proto-object learning 
    $\mathcal{L}_{\text{proto}}$ ;
    (3) teacher-filtered temporal consistency $\mathcal{L}_{\text{temp}}$.
    The teacher network $g_{\theta'}$ is updated using EMA. 
  }
  \label{fig:framework}
  \vspace{-8pt}
\end{figure}

\section{Methodology}
\vspace{-2pt}
\label{sec:methods}
Our goal is to learn persistent, class-agnostic object representations from unconstrained egocentric video.
We propose EgoViT, a teacher-student framework that bootstraps the learning by synergistically optimizing three core mechanisms: \textbf{1)} Proto-object Learning, \textbf{2)} Depth Regularization, and \textbf{3)} Teacher-Filtered Temporal Consistency, as illustrated in Figure~\ref{fig:framework}.
Our framework employs a student encoder $g_{\theta}$ and a momentum-updated teacher $g_{\theta'}$; the high-level training logic is outlined in Algorithm~\ref{alg:train-high-level}.
\subsection{Proto-object Delineation \& Learning}
\label{sec:proto_discovery_learning}
\begin{figure}[t]
    % \vspace{-19pt}
    \centering
    \includegraphics[width=\linewidth]{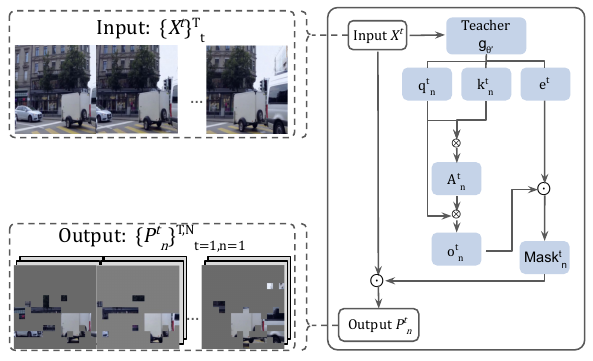}
    \vspace{-20pt}
    \caption{
    Proto-object Delineation via Teacher Attention.
    }
    \label{fig:proto}
\end{figure}

\begin{algorithm}[t]
    \caption{EgoViT: High-Level Training Logic}
    \label{alg:train-high-level}
    \small
    \begin{algorithmic}[1]
        \State \textbf{Input:} A batch of video clips $\{X^t\}$.
        \State \textbf{Output:} Student parameters $\theta$ and Teacher parameters $\theta'$.
        \State
        \For{\textit{For each training iteration:}}
        % \Comment{\textit{For each training iteration:}}
            \State Extract multi-level features and attention maps from the Student ($\theta$) and Teacher ($\theta'$) networks using the input clips $\{X^t\}$.
            \State Compute the intra-frame distillation losses ($\mathcal{L}_{\text{proto}}$).
            \State Compute the depth regularization loss $\mathcal{L}_{\text{depth}}$.
            \State Compute the teacher-filtered temporal consistency loss $\mathcal{L}_{\text{temp}}$.
            \State Aggregate the total loss $\mathcal{L}_{\text{total}}$ (See Eq.~\ref{eq:l_total}).
            \State Update $\theta$ via backpropagation.
            \State Update $\theta'$ via Exponential Moving Average (EMA).
        \EndFor
    \end{algorithmic}
\end{algorithm}

We now detail a key component that enables our model to move beyond patch-level representation learning to discover and represent coherent, object-like entities, which we term \emph{proto-objects}.
This section describes the two-stage process for each frame $t$: \textbf{1)} generating proto-object masks using the teacher's attention, and \textbf{2)} learning discriminative representations for these proto-objects via a dual-stream knowledge distillation framework.

\subsubsection{Proto-object Delineation via Teacher Attention}
\label{sec:proto-object_delineation}
For stable proto-object discovery, we utilize the momentum-updated teacher network. 
Proto-objects refer to latent, compositional, and temporally stable visual primitives that constitute complex scenes or complete objects.
Motivated by the discovery that attention heads can function as emergent object detectors\cite{caron2021emerging}, we encourage each of the $N$ attention heads in the teacher model's final layer to detect a distinct proto object.
To ensure this distinctiveness, we design an object consistency objective (\ref{sec:intra_frame_distillation}) and temporal consistency loss (\ref{sec:temporal_consistency}), which jointly enforce the learning of coherent and compositional proto-features.

As illustrated in Figure~\ref{fig:proto}, delineating the $n$-th proto-object at time $t$ involves three steps: prototype synthesis, spatial localization, and discrete mask generation.

\textbf{First}, we synthesize a head-specific prototype feature $\mathbf{o}^t_n$, which represents the pattern or concept that the $n$-th attention head is seeking in the current frame.
We define this prototype by aggregating the teacher network's query embedding $\mathbf{q}^t_n$ and its corresponding spatial attention map $\mathbf{A}^t_n$: $\mathbf{o}^t_n = \mathbf{A}^t_n \cdot \mathbf{q}^t_n$.

\textbf{Subsequently}, this prototype feature $\mathbf{o}^t_n$ is used to localize regions in the image that match the proto-object concept.
We compute a soft assignment map $\mathbf{M}^t_n$ via cosine similarity between the prototype $\mathbf{o}^t_n$ and every patch embedding $\mathbf{e}^t$ output by the teacher backbone: $\mathbf{M}^t_n = \text{sim}(\mathbf{o}^t_n, \mathbf{e}^t)$.
This map highlights image regions whose appearance matches the concept captured by the $n$-th head.

\textbf{Finally}, to obtain a discrete spatial mask for the proto-object, we binarize this soft assignment map using an adaptive thresholding strategy, $\text{Mask}^t_n = \mathbb{1}(\mathbf{M}^t_n > \mathbb{E}[\mathbf{M}^t_n])$.
This parameter-free approach dynamically segments the most salient regions corresponding to the prototype.

This generated mask is applied to the original input image $X^t$, producing a masked input $P^t_n = X^t \odot \text{Mask}^t_n$, thereby isolating the specific proto-object. 
This masked input $P^t_n$ is then fed into the student encoder $g_{\theta}$ to produce the corresponding \textbf{individual proto-object feature} $\mathbf{f}_n$.
This feature sets {$\{\mathbf{f}_n\}$} form the basis for the compositional consistency learning described in the next section.

\subsubsection{Proto-object Learning}
\label{sec:intra_frame_distillation}
Once proto-objects are delineated, we must ensure that the student network learns their high-quality feature representations. 
We achieve this through a knowledge distillation framework with a composite objective $\mathcal{L}_{\text{proto}}$.
This loss ensures consistency between two feature representations within the student network: the global feature $\mathbf{f^t}$ derived from the unmasked input and the compositional feature $\mathbf{f^t}_{\text{agg}}$ aggregated from the individual proto-object features, with the stable global target feature $\mathbf{f'}^t$ provided by the teacher network, as illustrated in Figure~\ref{fig:module}(b).

The alignment is measured by a cross-entropy loss, $H(y', y)$, between the softmax outputs of a teacher target $y'$  and a student prediction $y$, formulated as:
\vspace{-1pt}
{\small
\begin{equation}
H(y', y) = - \! \sum_{i=1}^{C} \text{softmax}\! \left(\frac{y'}{\tau_t}\right)_i \!\! \cdot \text{log\_softmax}\! \left(\frac{y}{\tau_s}\right)_i
\label{eq:h_definition}
\vspace{-2pt}
\end{equation}
}
where $C$ is the output dimensionality; $\tau_t$ and $\tau_s$ are the teacher and student temperatures, respectively, which control the sharpness of their output distributions.

Our composite loss, $\mathcal{L}_{\text{proto}}$, sums two alignment terms:
\vspace{-1pt}
{\small
\begin{equation}
\mathcal{L}_{\text{proto}} = H(\mathbf{f'}, \mathbf{f}) + H(\mathbf{f'}, \mathbf{f}_{\text{agg}})
\label{eq:l_proto}
\vspace{-2pt}
\end{equation}
}
The \textbf{first} term, $H(\mathbf{f'}, \mathbf{f})$, ensures holistic scene understanding by aligning the student's global feature $\mathbf{f}$ from the unmasked input.
% to teacher's feature $\mathbf{f}'$. 
The \textbf{second} term, $H(\mathbf{f'}, \mathbf{f}_{\text{agg}})$ grounds the learning in object-level entities by aligning a compositional feature, $\mathbf{f}_{\text{agg}}$, which is a weighted average of the student's individual proto-object features (see Appendix~\ref{appendix:method-details} for details). 
This composite objective teaches the model to build a holistic understanding from its constituent parts.

\subsection{Depth Regularization}
\label{sec:geometric_priors}
While proto-object learning provides discriminative appearance features, these cues alone are insufficient to stably decouple objects from the background amidst the continuous ego-motion and cluttered scenes of egocentric video. 
Moreover, traditional motion cues like optical flow are computationally expensive and often unavailable as direct sensor inputs. 
Inspired by how primate visual systems rely on geometric perception for stable world modeling~\cite{wolbers2008spatial,allman1999evolving}, we introduce an auxiliary depth regularization task. 
This grounds our representations in physical reality using a geometric cue that, unlike optical flow, is readily available from sensors.

\begin{figure*}[t!]
    \centering
    \vspace{-8pt}
    \includegraphics[width=\linewidth]{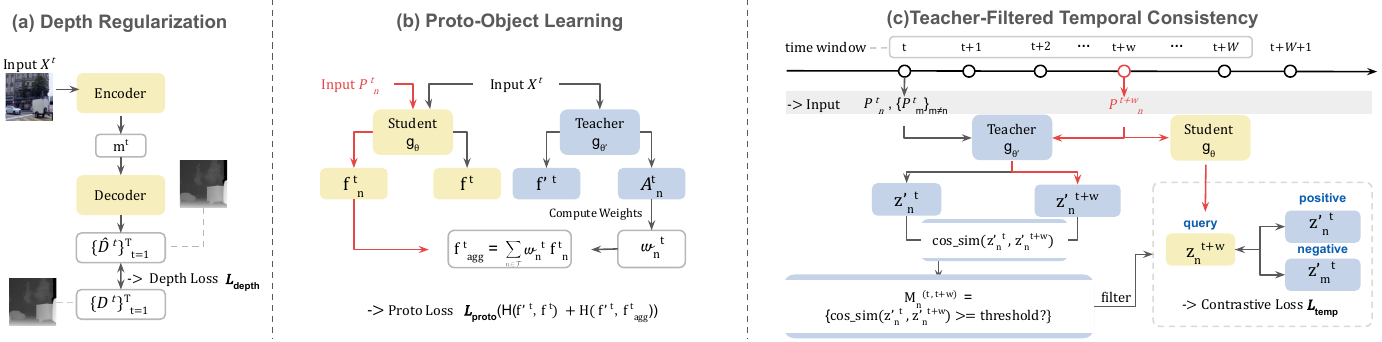}
    \vspace{-20pt}
    \caption{
        % The EgoViT framework.
        \textbf{(a) Depth Regularization:} An auxiliary task, $\mathcal{L}_\text{depth}$, provides geometric constraint.
        \textbf{(b) Proto-Object Learning:} A distillation loss, $\mathcal{L}_\text{proto}$, aligns student and teacher features in proto-level. Here, $H(y', y)$ denotes the cross-entropy between the softmax outputs of a teacher target $y'$ and a student prediction $y$.
        \textbf{(c) Teacher-Filtered Temporal Consistency:} A contrastive loss $\mathcal{L}_\text{temp}$ is applied on reliable pairs filtered by the teacher to enforce temporal identity.
    }
    \label{fig:module}
    \vspace{-5pt}
\end{figure*}

To achieve this, we guide the learning process with a structured loss, $\mathcal{L}_{\text{depth}}$ (detailed in Appendix~\ref{appendix:method-details}). This loss consists of a scale-invariant term for capturing relative layouts and a gradient consistency term for preserving object boundaries.
This design allows the optimization process to focus on geometric structure itself while ignoring unreliable absolute scale and shift in depth, thereby effectively regularizing our self-supervised representations.
As illustrated in Figure~\ref{fig:module}(a), intermediate features $m^t$ from the student encoder are fed into a lightweight decoder (architecture detailed in Appendix~\ref{appendix:depth_decoder}).
Notably, this geometric regularization task is performed only during training and requires no geometric input during inference.

\subsection{Teacher-Filtered Temporal Consistency}
\label{sec:temporal_consistency}
While the previous objective (Eq.\ref{eq:l_proto}) teaches the model to identify proto-objects within a single frame, a critical challenge remains: ensuring these representations are consistent over time, especially through occlusions and rapid ego-motion. 
To address this, we introduce a Teacher-Filtered Temporal Consistency mechanism.

Our innovation is to leverage the stable, momentum-updated teacher to proactively filter out unreliable temporal correspondences before they can corrupt the student's training signal. 
The process for any two frames, $t$ and $t'$ within a temporal window $W$ ($|t - t'| \le W$), contains two stages:

\vspace{-5pt}
\paragraph{Filtering via Teacher Agreement.}
First, we assess the correspondence reliability for the $n$-th proto-object by computing the cosine similarity between their respective feature representations from the \textit{teacher} network, $\mathbf{z}'^t_n$ and $\mathbf{z}'^{t'}_n$. 
This score quantifies the temporal coherence as perceived by the stable teacher.
We then generate a dynamic binary mask, $M^{(t, t')}_n$, by thresholding this similarity score, defined as $M^{(t, t')}_n = \mathbb{1}\left( \text{sim}(\mathbf{z}'^t_n, \mathbf{z}'^{t'}_n) > \lambda \right)$.
Here, $\text{sim}(\cdot, \cdot)$ is the cosine similarity and $\lambda$ is a confidence threshold (e.g., $\lambda=0.8$. We found this threshold to be robust across datasets. A sensitivity analysis for lambda is provided in the Appendix~\ref{appendix:additional-experimental-results}).
This mask effectively prunes unreliable pairs where the object might have been occluded, left the frame, or drastically changed in appearance.

\vspace{-5pt}
\paragraph{Temporal Contrastive Objective.}
Second, we apply a temporal contrastive loss only on these filtered pairs.
The loss encourages the \textbf{student's} representation of a proto-object at time $t'$, $\mathbf{z}^{t'}_n$, to be close to the corresponding \textbf{teacher's} representation at time $t$, $\mathbf{z}'^t_n$ (the positive pair), while being dissimilar to all other proto-object representations from the teacher at time $t$ (the negative pairs).
Here, the feature $\mathbf{z}_n$ is the penultimate layer's output (i.e., the bottleneck feature) from the student's projection head.
The loss for a valid pair $(t, t')$ is:
{\small
\begin{equation}
\mathcal{L}^{(t, t')}_{\text{temp}} = -\frac{1}{|\mathcal{P}|} \sum_{n \in \mathcal{P}} \log \frac{
\exp(\text{sim}(\mathbf{z}^{t'}_n, \mathbf{z}'^t_n) / \alpha)}
{\sum_{k=1}^{\mathcal{K}} \exp(\text{sim}(\mathbf{z}^{t'}_n, \mathbf{z}'^t_k) / \alpha)}
\label{eq:l_temp}
\end{equation}
}
where $\mathcal{P} = \{n \mid M^{(t, t')}_n = 1\}$ is the set of valid proto-objects, $\mathcal{K}$ is the total number of proto-objects considered, and $\alpha$ is a temperature parameter. 
The total temporal loss $\mathcal{L}_{\text{temp}}$ is the average over all such temporal pairs in the clip.

\subsection{Joint Objective and Emergent Synergy.}
Finally, the proto-object learning ($\mathcal{L}_{\text{proto}}$), depth regularization ($\mathcal{L}_{\text{depth}}$), and temporal consistency ($\mathcal{L}_{\text{temp}}$) are optimized jointly via a single, weighted objective:
{\small
\begin{equation}
\mathcal{L}_{\text{total}} = \gamma_{P} \mathcal{L}_{\text{proto}} + \gamma_{D} \mathcal{L}_{\text{depth}} + \gamma_{T} \mathcal{L}_{\text{temp}}
\label{eq:l_total}
\end{equation}
}
where $\gamma_{P}$,$\gamma_{D}$ and $\gamma_{T}$ are hyperparameters used to balance the magnitudes of the individual loss components.
We set them to $0.3$, $1.0$, and $0.5$ respectively in our experiments.
We find this joint optimization is critical for learning robust, persistent object representations from unconstrained egocentric video.

\begin{table*}[t]
    \centering
    \scriptsize
    \caption{
    Comparison of EgoViT with prior methods on downstream tasks.
     Our default model, \textbf{EgoViT$_{\text{Zurich}}$} is trained on a single 65-minute video, while \textbf{EgoViT$_{\text{WT-all}}$} is trained on the full Walking Tours dataset to evaluate scalability.
    \textbf{Bold} indicates the best result in each column.
    Numbers in parentheses (\textcolor[rgb]{0.2,0.6,0.2}{green}) denote absolute gains of \textbf{EgoViT$_{\text{Zurich}}$} over DINO.
    All metrics are reported in \%, and higher is better.
    }
    \vspace{-10pt}
    \begin{tabular}{l|cc|c|c|c|c|c|c|c|c}
    \hline
    \multirow{3}{*}{METHOD} & \multicolumn{2}{c|}{Semantic SEG.} & \multicolumn{1}{c|}{Object DET.} & \multicolumn{1}{c|}{Instance SEG.} & \multicolumn{3}{c|}{Video Object Segmentation} &  \multicolumn{1}{c|}{Object DIS.}& \multicolumn{2}{c}{Classification} \\
    \cline{2-11}
    & mIoU & Acc$_m$ & mAP  & mAP & $(\mathcal{J}\&\mathcal{F})_m$ & $\mathcal{J}_m$ & $\mathcal{F}_m$  & CORLOC & LP & $k$-NN \\
    \hline
    SimCLR~\cite{chen2020simple} & 22.5 & 32.8 & 22.2 & 20.1 & 52.3 & 51.5 & 53.1 & 35.5 & 28.2 & 33.1 \\
    AttMask~\cite{kakogeorgiou2022attmask}  & 25.1 & 35.4 & 25.9 & 23.9 & 52.2 & 52.5 & 51.8 & 37.8 & 25.4 & 35.9 \\
    MoCo-v3~\cite{chen2021empirical}  & 17.9 & 26.0 & 19.0 & 17.3 & 52.5 & 50.9 & 54.0 & 43.1 & 22.5 & 31.2  \\
    MAE~\cite{he2022masked}  & 23.0 & 33.2 & 24.6 & 22.1 & 51.3 & 50.3 & 52.2 & 35.9 & 13.0 & 17.8  \\
    iBOT~\cite{zhou2021ibot}  & 23.9 & 34.5 & 22.1 & 19.6 & 53.9 & 53.5 & 54.3 & 36.3 & 27.5 & 33.6 \\
    DORA~\cite{venkataramanan2023imagenet}  & 21.6 & 31.1 & 22.6 & 20.4 & 53.8 & 51.9 & 55.6 & 24.1 & 29.6 & 33.7  \\
    \hline
    DINO~\cite{caron2021emerging}  & 21.2 & 30.3 & 22.0 & 20.6 & 53.8 & 52.5 & 55.1 & 37.2 & 30.9 & 35.5  \\
    EgoViT$_{\text{Zurich}}$ & 26.0(\textcolor[rgb]{0.2,0.6,0.2}{+4.8}) & 36.6(\textcolor[rgb]{0.2,0.6,0.2}{+6.3}) & 26.7(\textcolor[rgb]{0.2,0.6,0.2}{+4.7}) & 24.3(\textcolor[rgb]{0.2,0.6,0.2}{+3.7}) & 54.3(\textcolor[rgb]{0.2,0.6,0.2}{\scriptsize +0.5}) & 52.7(\textcolor[rgb]{0.2,0.6,0.2}{\scriptsize +0.2}) & 55.9(\textcolor[rgb]{0.2,0.6,0.2}{\scriptsize +0.8}) & 45.2(\textcolor[rgb]{0.2,0.6,0.2}{\scriptsize +8.0}) & 34.0 (\textcolor[rgb]{0.2,0.6,0.2}{\scriptsize +3.1}) & 38.9(\textcolor[rgb]{0.2,0.6,0.2}{\scriptsize +3.4}) \\
    EgoViT$_{\text{WT-all}}$ & \textbf{30.6} & \textbf{39.3}  & \textbf{29.6}  & \textbf{26.8} & \textbf{57.0}  & \textbf{55.0} & \textbf{58.9}  & \textbf{50.2} & \textbf{39.1}  & \textbf{45.3} \\
    \hline
    \end{tabular}
    \label{tab:main_results}
\end{table*}  

\section{Experiments}
\vspace{-1pt}
\subsection{Experimental Setup}
\vspace{-1pt}
\paragraph{Pre-training Data and Baselines.}
Our framework, EgoViT, is designed for a self-supervised paradigm grounded in egocentric video, where appearance (RGB) and geometric (depth) data streams can be captured directly by hardwares like RGB-D camera, without human annotation.
For our primary experiments, we leverage a single, unconstrained egocentric 65-minute long video WT-Zurich from the Walking Tours (WT) dataset~\cite{venkataramanan2023imagenet} as our main training corpus, and denote the trained model as EgoViT$_{\text{Zurich}}$.
As WT-Zurich is RGB-only, we simulate the availability of a depth sensor by generating a pseudo-depth channel using the off-the-shelf monocular estimator, Depth-Anything-V2~\cite{yang2024depth}. 
We further verify in Sec. 4.3 that EgoViT remains robust across depth priors of varying quality, underscoring that it learns from the underlying geometric structure rather than the specifics of a pre-trained estimator.

We compare our main model, EgoViT$_{\text{Zurich}}$, against a comprehensive suite of state-of-the-art self-supervised methods from both image (e.g., DINO~\cite{caron2021emerging}, iBOT~\cite{zhou2021ibot}, MAE~\cite{he2022masked}) and video (e.g., DoRA~\cite{venkataramanan2023imagenet}) domains. 
In addition, we also discuss the comparison of different design goals (e.g., segmentation-oriented models such as Poodle~\cite{wang2024poodle}, SAVi++~\cite{elsayed2022savi++}) in the appendix~\ref{appendix:Additional-Experimental-Results}.

\vspace{-5pt}
\paragraph{Note on Baseline Reproducibility}
To ensure the fairest possible comparison, all baseline models were re-implemented and evaluated within our unified experimental framework. 
This process guarantees that all methods were subject to the exact same data processing, training schedule, and evaluation protocol. 
For full transparency, all baseline results reported in this paper are generated from this controlled re-implementation. 
Further re-implementation details are available in the Appendix~\ref{appendix:implementation}.

\vspace{-5pt}
\paragraph{Implementation Details.}
Our EgoViT framework employs a ViT-Small (ViT-S/16) backbone, initialized from scratch. 
We train all models using the AdamW optimizer with an effective batch size of 192. 
The learning rate starts at a base of 5e-4, warms up for 10 epochs, and then decays following a cosine schedule to a minimum of 1e-5. 
We employ a weight decay strategy and is linearly increased from 0.04 to 0.4 over the course of training. 
Our main models are trained for 320 epochs, while ablation studies are conducted for an efficient 40 epochs. All experiments use a fixed random seed for reproducibility. Full hyperparameter details are in Appendix~\ref{appendix:implementation}.
Code will be made publicly available.

\vspace{-5pt}
\paragraph{Downstream Tasks and Evaluation.}
We evaluate our learned representations on a diverse suite of downstream tasks.
These include: 1) semantic segmentation on ADE20K~\cite{zhou2017scene}, 2) object detection and instance segmentation on Mini MS-COCO~\cite{samet2020houghnet}, 3) video object segmentation on DAVIS-2017~\cite{pont20172017}, 4) unsupervised object discovery on PASCAL VOC 2012~\cite{pascal-voc-2012}, and 5) image classification on ImageNet-1k~\cite{deng2009imagenet}.
We follow standard protocols for all tasks, with detailed settings provided in Appendix~\ref{appendix:implementation}.
In addition, we further evaluate generalization on Ego4D~\cite{grauman2022ego4d}, with the corresponding results provided in Appendix~\ref{appendix:ego4d}. Comparisons of several models on WT-Venice are presented in Appendix~\ref{appendix:additional-experimental-results}.

\subsection{Main Results and Analysis}
Table~\ref{tab:main_results} compares the proposed \textbf{EgoViT$_{\text{Zurich}}$} against state-of-the-art methods across a range of downstream tasks.
Note that for direct and fair comparisons, all models, including the baselines, are pre-trained from scratch on the same 65-minute WT-Zurich video under our unified protocol.
Obviously, EgoViT demonstrates significant improvements, particularly on tasks requiring robust object-level understanding.
Key findings are detailed below.
Appendix~\ref{appendix:implementation} provides additional generalization results on other datasets.
EgoViT$_{\text{WT-all}}$ will be discussed in Section \ref{paragraph:dataset all}.

\vspace{-5pt}
\paragraph{Strong Performance on Dense Prediction Tasks.}
The geometric priors learned by EgoViT provide a powerful foundation for dense prediction tasks. This is evidenced by its performance on ADE20k semantic segmentation, where EgoViT$_{\text{Zurich}}$ outperforms DINO by a remarkable +4.8\% mIoU. We attribute this substantial gain directly to our depth regularization module (D), which instills a structural awareness in the learned features, facilitating more accurate figure-ground distinction. This advantage is further confirmed in instance-level tasks, with EgoViT$_{\text{Zurich}}$ achieving a +3.7\% mAP gain on MS-COCO instance segmentation.

\vspace{-5pt}
\paragraph{Superior Video Object Segmentation and Generalization.}
EgoViT demonstrates powerful generalization by successfully transferring its learned representations to the task of video object segmentation. 
On the DAVIS-2017 benchmark, which consists of third-person videos with unseen object categories, our model achieves a $(\mathcal{J\& F})_m$ score of 54.3\%. 
Its performance on contour precision is strong ($\mathcal{F}_m$ at 55.9\%), which we attribute to our temporal alignment mechanism (T) promoting consistent boundary tracking through occlusions.
This result validates that features learned from a single egocentric video are robust and transferable to conventional video understanding tasks.
We further validate this on the egocentric Epic-Kitchens VISOR benchmark in Appendix~\ref{appendix:Additional-Experimental-Results}.

\vspace{-5pt}
\paragraph{Dominance in Unsupervised Object Discovery.}
EgoViT shows exceptional performance in unsupervised object discovery.
On the PASCAL VOC benchmark, EgoViT$_{\text{Zurich}}$ achieves 45.2\% CorLoc, a significant +8.0\% improvement over the DINO baseline. 
This result highlights the effectiveness of our model's design, where the combination of proto-object learning (P) and temporal consistency (T) enables the discovery and persistent tracking of objects, a key capability for open-world localization, as further illustrated zero-shot on Ego4D in Appendix ~\ref{appendix:ego4d}.

\vspace{-5pt}
\paragraph{Effective Transfer to Image Classification.}
Finally, the features learned from complex egocentric dynamics transfer effectively to standard image classification. On ImageNet-1k, our model achieves up to 34.0\% linear probing accuracy, demonstrating that the learned representations are versatile and not limited to object-centric tasks.

\vspace{-5pt}
\paragraph{Scalability on Pre-training Data.}
\label{paragraph:dataset all}
To evaluate the scalability of EgoViT on training data, we trained an additional model, \textbf{EgoViT$_{\text{WT-all}}$}, on the entire \textit{Walking Tours} dataset. 
The final row of Table~\ref{tab:main_results} shows that access to substantially more diverse data provides a significant and consistent performance boost across all tasks (e.g., CorLoc improves from 45.2\% to 50.2\%, and SemanticSeg mIoU increases from 26.0\% to 30.6\%). 
This confirms that while our method is highly effective when trained on a single video, it also scales gracefully with increasing data volume and diversity, highlighting a promising direction for future work.
In addition, we provide comprehensive analysis of how training video length affects EgoViT's performance in Appendix~\ref{appendix:Additional-Experimental-Results}.

\begin{table}[t]
\centering
\scriptsize
\caption{
    Component ablation study. The results highlight the individual contributions of D, P, and T. 
    Notably, combining P and T without D (P+T) leads to performance degradation, underscoring the importance of geometric prior.
}
\vspace{-8pt}
\begin{tabular}{ ccc cc}
\toprule
\cmidrule(r){1-4} % booktabs提供的局部橫線，更美觀
D & P & T & $k$-NN & CORLOC\\
\midrule  
$\times$ & $\times$ & $\times$ & 21.8 & 27.5 \\
\midrule
%\multicolumn{4}{l}{\textit{Single Components}} \\
$\checkmark$ & $\times$ & $\times$ & 22.2 & 34.6 \\
$\times$ & $\checkmark$ & $\times$ & 22.0 & 35.6 \\
$\times$ & $\times$ & $\checkmark$ & 22.2 & 35.2 \\
\midrule
%\multicolumn{4}{l}{\textit{Two-Component Synergy}} \\
$\checkmark$ & $\checkmark$ & $\times$ & 22.5 & 36.2 \\
%\rowcolor{gray!15}
$\checkmark$ & $\times$ & $\checkmark$ & \textbf{22.9} & 37.9 \\
%\rowcolor{gray!15}
$\times$ & $\checkmark$ & $\checkmark$ & 22.4 & 35.9 \\
\midrule
$\checkmark$ & $\checkmark$ & $\checkmark$ & \textbf{23.2} & \textbf{38.3}\\
\bottomrule
\end{tabular}
\label{tab:ablation_components}
\end{table}

\begin{table}[t]
\centering
\scriptsize
\caption{
    Ablation on temporal modeling strategies. Our full proto-level approach (with the teacher-filtered) is the only strategy that significantly outperforms the baseline, highlighting the filter's essential role.
}
\vspace{-8pt}
\begin{tabular}{lcccc}
\toprule
Method & Strategy & Frames & $k$-NN & CORLOC \\
\midrule
EgoViT (D+T) & Frame & 3 & 21.5 $\downarrow$ & 34.2 $\downarrow$ \\
% \rowcolor{gray!15}
EgoViT (D+T$_\text{w/o filter}$) & Proto & 3 & 21.5 & 31.4 $\downarrow$ \\
% \rowcolor{gray!15} 
EgoViT (D+T$_\text{w/o filter}$) & Proto & 4 & 21.2 & 33.2 $\downarrow$ \\
\midrule
% --- Our Full Approach ---
{EgoViT (D+T) } & {Proto} & 3 & 22.7 & 35.9 $\uparrow$ \\
{EgoViT (D+T) } & {Proto} & 4 & \textbf{22.9} & \textbf{37.9} $\uparrow$ \\
{EgoViT (D+T) } & {Proto} & 5 & 22.0 & 35.4 $\uparrow$ \\
\bottomrule
\end{tabular}
%}
\label{tab:ablation_strategy}
\end{table}

\subsection{Ablation Studies}
\vspace{-3pt}
We conduct a series of ablation studies on two representative downstream tasks: $k$-NN classification on ImageNet-1k and unsupervised object discovery (CorLoc) on PASCAL VOC 2012~\cite{pascal-voc-2012}. 
We present our main ablation analyses below.
Additional qualitative analyses of the learned components and hyperparameter sensitivity studies are provided in Appendix~\ref{appendix:Ablation-Study-Details}.

\vspace{-5pt}
\paragraph{Component Contribution: Synergy of Geometry and Time.}
Our primary ablation study in Table~\ref{tab:ablation_components} dissects the contributions of our core components, Depth (D), Proto-objects (P), and Temporal learning (T), and reveals a crucial insight for egocentric object discovery: the emergence of stable representations is driven by the synergy between geometric grounding and temporal reasoning.

First, we establish that explicitly modeling object-like structure is the primary driver of performance. 
Individually, both the depth-guided module (D) and the proto-object learning module (P) dramatically outperform the DINO baseline in object discovery, confirming the value of moving beyond simple patch-level correspondence.

The most compelling finding, however, lies in their interaction with the temporal learning module (T). 
While combining depth and time (D+T) yields the best two-component performance (37.9\% CorLoc), the proto-object and time (P+T) combination leads to only a limited performance improvement at 35.9\% CorLoc.
This result strongly suggests that without the stable geometric grounding provided by depth, temporal learning can be misled by the ambiguous appearance of proto-objects, especially under the severe ego-motion of first-person video.

Ultimately, the full EgoViT model (D+P+T) achieves the highest performance, confirming that all three components are complementary.
In essence, depth provides the stable geometric context of \textbf{where} an object is, proto-objects provide the initial grouping of \textbf{what} might be an object, and temporal learning tracks \textbf{how} that object persists over time.

\vspace{-5pt}
\paragraph{Temporal modeling strategies.}
Table \ref{tab:ablation_strategy} validates the effectiveness of our teacher-filtered, proto-level temporal consistency strategy. We find that unfiltered temporal matching is ineffective; a naive frame-level strategy slightly hurts performance.
A proto-level strategy without our filter (EgoViT (D+T$_\text{w/o filter}$)) causes a significant drop in CorLoc from the baseline EgoViT (D+T) of 37.9\% to 33.2\% when Frames=4. This highlights the challenge of noisy correspondences in egocentric video.

On the other hand, a Frame-level strategy causes a significant drop in CorLoc from the baseline of 35.9\% to 34.2\% when Frames=3. 
These comparisons underscores that our filtering mechanism and the proto-level strategy are the key components that unlocks effective temporal self-supervision from unconstrained video streams.

\begin{table}[t]
\centering
\scriptsize
\caption{
Robustness to depth quality and source. We retrain EgoViT with (a) Gaussian blurred depth ($\sigma=\sigma_{0}\times W/224$, scaled to image size) and (b) alternative monocular estimators (MiDaS, Depth-Pro). 
}
\vspace{-6pt}
\begin{tabular}{lcc}
\toprule
Method & $k$-NN & CORLOC \\
\midrule
w/o Depth (no D) & $22.4$ & $35.9$ \\
Gaussian blur ($\sigma_{0}=0.15$) & $22.9$ & $37.7$ \\
Gaussian blur ($\sigma_{0}=0.3$)  & $22.7$ & $37.2$ \\
Gaussian blur ($\sigma_{0}=0.6$)  & $22.6$ & $36.4$ \\
\midrule
MiDaS (other estimator) & $22.6$ & $36.7$ \\
% \rowcolor{gray!15}
Depth-Pro (other estimator) & $\mathbf{23.5}$ & $\mathbf{38.6}$ \\
Full (original depth) & $23.2$ & $38.3$ \\
\bottomrule
\end{tabular}
\label{tab:robust_depth_quality_source}
\end{table}

\vspace{-5pt}
\paragraph{Robustness to Depth Quality and Source.}
\label{paragraph:depth}
To assess the dependence of EgoViT on geometric accuracy, we retrain the model using (a) Gaussian-blurred depth maps and (b) alternative monocular estimators (Table~\ref{tab:robust_depth_quality_source}).
We observe that performance degrades only modestly as the depth becomes increasingly blurred, indicating that the model mainly relies on coarse structural cues rather than precise geometry.
When replacing the original depth with off-the-shelf estimators such as MiDaS~\cite{ranftl2020towards} or Depth-Pro~\cite{bochkovskii2024depth}, EgoViT maintains comparable performance; the Depth-Pro variant performs slightly better on average.
These results suggest that our geometric branch is robust to both depth quality and source, which may broaden the applicability of EgoViT in real-world settings where accurate depth is not available.

\subsection{Temporal Stability and Long-Term Tracking}
\begin{figure}[t]
    \centering
    \includegraphics[width=\linewidth, keepaspectratio]{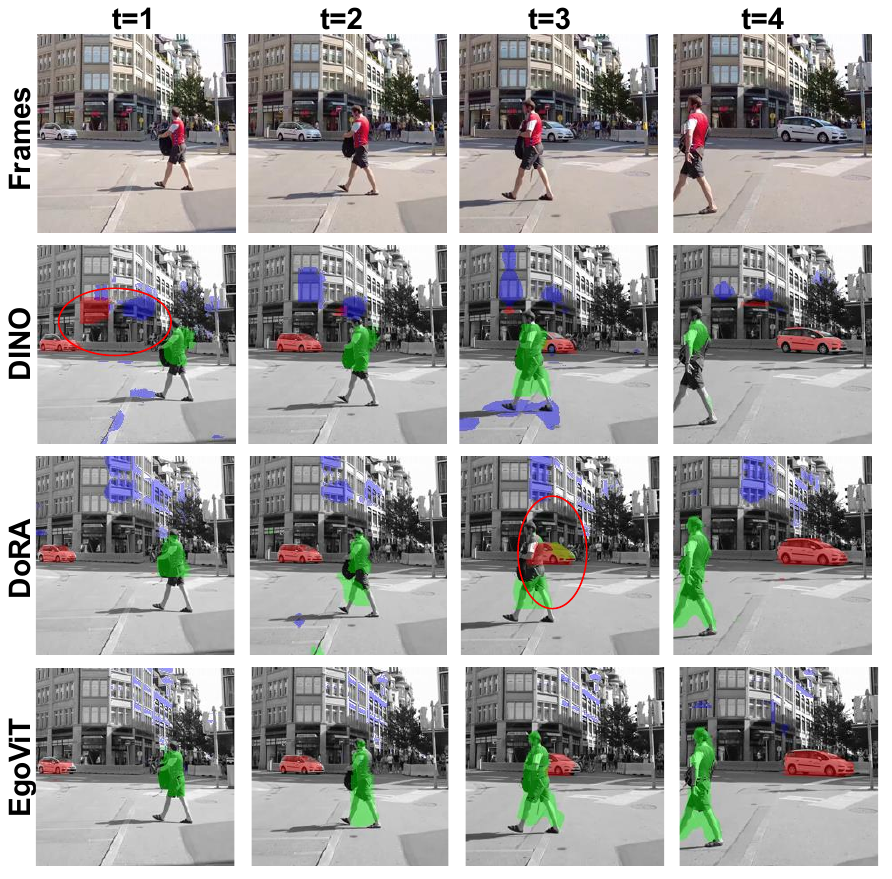}
    \vspace{-20pt}
    \caption{
        EgoViT achieves superior temporal attention stability.
        Compared to baselines (DINO, DoRA) that exhibit significant attention drift over time, our method maintains a coherent focus on the target object across the sequence, even through severe occlusion (see red circles for failure cases).
    }
    \label{fig:visualization}
\end{figure}

\vspace{-1pt}
\paragraph{Qualitative Results.}
Figure~\ref{fig:visualization} provides compelling visual evidence of EgoViT's temporal stability compared to prior work.
The attention maps of DINO and DoRA exhibit significant drift, frequently losing track of the target object to focus on background structures (DINO at $t=1,3$) or failing to re-associate with the object after occlusion (DoRA at $t=3$).
In stark contrast, EgoViT, explicitly guided by its temporal consistency objective (T), maintains a stable and coherent head-to-object association across the entire sequence, demonstrating its robustness to severe viewpoint changes and occlusions.
We provide more qualitative examples across diverse scenarios in Appendix~\ref{appendix:more_visualization}.

\vspace{-6pt}
\paragraph{Quantitative Results on LaSOT.}
To further assess long-term consistency under challenging conditions, we evaluate self-supervised backbones on the LaSOT benchmark \cite{fan2021lasot}, which features long sequences with frequent occlusions and multiple disappearance–reappearance events.
Following standard practice, we adopt the OSTrack framework \cite{ye2022joint} and keep its tracking head, hyperparameters, and evaluation protocol unchanged across all backbones to ensure a fair comparison.
As shown in Table~\ref{tab:dynamic_tracking_lasot}, EgoViT consistently outperforms both DINO and DoRA under the same settings.
Additional tracking results on other benchmarks are reported in the Appendix~\ref{appendix:Additional-Experimental-Results}.

\begin{figure}[t]
    \centering
    \includegraphics[width=\linewidth]{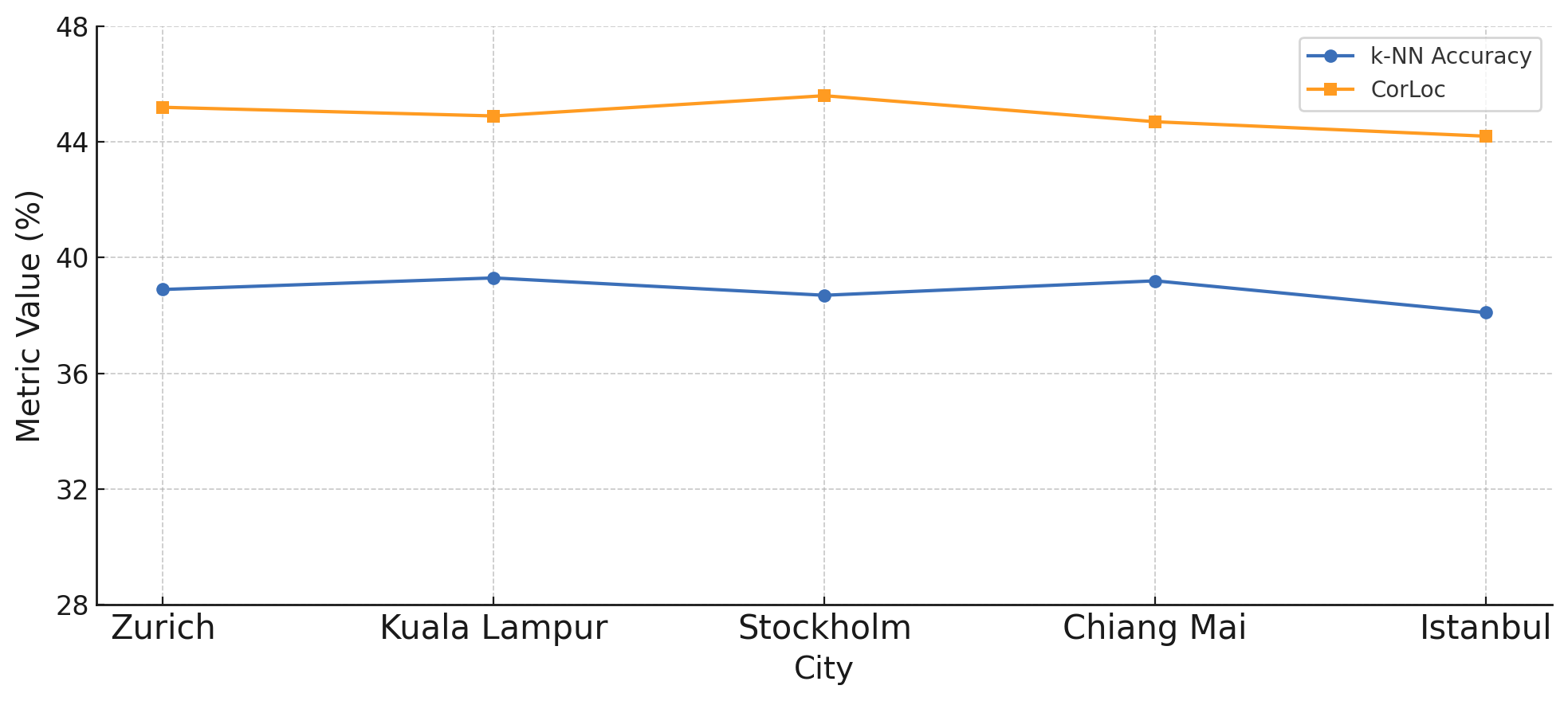}
    \vspace{-20pt}
    \caption{
        EgoViT demonstrates robust generalization across diverse visual environments.
    }
    \label{fig:egovit_lineplot}
    \vspace{-5pt}
\end{figure}

\vspace{-6pt}
\paragraph{Stability across Visual Environments.}
To examine the stability of EgoViT when trained on videos of comparable duration, we evaluate the model using multiple $\sim$60-minute egocentric videos collected from five visually diverse cities (Zurich, Istanbul, Stockholm, Chiang Mai, and Kuala Lumpur).
Notably, these videos are not all captured under favorable lighting conditions; some contain extensive dusk or nighttime segments.
As shown in Figure~\ref{fig:egovit_lineplot}, EgoViT delivers consistently strong performance across all locations on both $k$-NN accuracy and CorLoc, with only minor variation despite differences in appearance, illumination, and motion dynamics.
Overall, these results demonstrate that EgoViT generalizes robustly across diverse real-world environments even when trained on videos of similar length.

\begin{table}[t]
\centering
\scriptsize
\caption{
Dynamic Tracking on LaSOT. Evaluated under long sequences with occlusion and target disappearance–reappearance using standard metrics (AUC, $P$, $P_{\text{norm}}$). 
}
\vspace{-6pt}
\begin{tabular}{lccc}
\toprule
Backbone & AUC & P & $P_{\text{norm}}$ \\
\midrule
DINO  & 60.5 & 64.7 & 70.1 \\
DoRA  & 61.7 & 66.4 & 72.3 \\
\rowcolor{gray!15}
EgoViT & \textbf{64.7} & \textbf{70.6} & \textbf{74.1} \\
\bottomrule
\end{tabular}
\label{tab:dynamic_tracking_lasot}
\vspace{-6pt}
\end{table}

\section{Conclusion and Future work}
\vspace{-1pt}
We introduced EgoViT, a novel framework that learns persistent object representations from unconstrained egocentric video.
We demonstrate with extensive experiments that, under challenging protocol, EgoViT significantly outperforms strong self-supervised baselines across a diverse suite of downstream tasks. 
Our work represents a significant step from static recognition towards the dynamic modeling of object permanence,offering a valuable foundation for the next generation of world models in embodied AI.

This work also highlights several promising directions for further exploration.
On one hand, incorporating semantic cues from large language models may help resolve early-stage visual ambiguities, thereby improving the stability of emerging object structure.
On the other hand, richer multi-view inputs open up new possibilities for modeling object persistence.
Collectively, these directions represent important steps toward scaling the framework to support lifelong learning agents in open-world environments.

\section*{Acknowledgments}
    This work is supported in part by the National Natural Science Foundation of China (No. 62206259), in part by the Fundamental Research Funds for the Central Universities (No. CUC25CGJ02). The authors would also like to acknowledge the VATIS Key Laboratory, Ministry of Culture and Tourism.

% \clearpage
{
    \small
    \bibliographystyle{ieeenat_fullname}
    \bibliography{main}
}

\clearpage
\clearpage
\setcounter{page}{1}
\maketitlesupplementary

\setcounter{section}{0}
\renewcommand{\thesection}{\Alph{section}}
\appendix

\noindent This appendix provides supplementary materials for the main paper. The contents are organized as follows:
\begin{itemize}
    \item Detailed Method Description (Sec.~\ref{appendix:method-details})
    \item Depth Decoder Architecture (Sec.~\ref{appendix:depth_decoder})
    \item Experimental Details (Sec.~\ref{appendix:implementation})
    \item On the Reproducibility of the DoRA Baseline (Sec.~\ref{appendix:reproducibility})
    \item Additional Experimental Results (Sec.~\ref{appendix:additional-experimental-results})
    \item More Visualization (Sec.~\ref{appendix:more_visualization})
    \item Broader Impacts and Ethical Considerations (Sec.~\ref{appendix:broader-impacts})
    
    \item Detailed Comparison and Positioning Analysis with DINO and DoRA (Sec.~\ref{app:detailed Comparison})
    \item Zero-Shot Generalization: Ego4D Case Study
    (Sec.~\ref{appendix:ego4d})
\end{itemize}

\section{Detailed Method Description}
\label{appendix:method-details}

\noindent
We provide the notation used throughout the paper in Table \ref{tab:notation}, and a more detailed description of our method components in the following part.

\begin{table*}[ht]
 \caption{\textbf{Notation used throughout the paper.} All variables are indexed by frame time $t$. For shapes, $S$ denotes the number of spatial patch tokens. $D_1, D_2, D_3$ are layer-specific feature dimensions for the backbone. $D, D_q, D_k$ are the feature dimensions for patch embeddings and attention components.}
 \label{tab:notation}
 \centering
 \small
\setlength{\tabcolsep}{6pt}
 \begin{tabular}{llll}
    \toprule
    \textbf{Symbol} & \textbf{Meaning} & \textbf{Shape / Dim.} & \textbf{Source} \\
    \midrule
    $X^{t}$ & RGB input frame & $H \times W \times 3$ & Input \\
    $P^{t}_{n}$ & Masked RGB input for proto-object $n$ & $H \times W \times 3$ & Input \\
    \midrule
    \multicolumn{4}{l}{\textit{Student Network Outputs}} \\
 $\mathbf{m}^{t}$ & Middle-layer representation & $\mathbb{R}^{D_1}$ & Student \\
 $\mathbf{z}^{t}$ & Penultimate-layer feature & $\mathbb{R}^{D_2}$ & Student \\
 $\mathbf{f}^{t}$ & Final-layer feature (unmasked input) & $\mathbb{R}^{D_3}$ & Student \\
 $\mathbf{f}^{t}_{n}$ & Individual proto-object feature & $\mathbb{R}^{D_3}$ & Student \\
    $\mathbf{f}^{t}_{\text{agg}}$ & Aggregated compositional feature & $\mathbb{R}^{D_3}$ & Student \\
    $\hat{D}^{t}$ & Predicted depth map & $H \times W$ & Student \\
    \midrule
    \multicolumn{4}{l}{\textit{Teacher Network Outputs \& Pseudo-Labels}} \\
    ${D}^{t}$ & Pseudo-ground truth depth map & $H \times W$ & frozen backbone~\cite{yang2024depth} \\
    $\mathbf{z}^{\prime t}$ & Penultimate-layer feature & $\mathbb{R}^{D_2}$ & Teacher \\
    $\mathbf{f}^{\prime t}$ & Final-layer feature & $\mathbb{R}^{D_3}$ & Teacher \\
    $\mathbf{e}^{t}$ & Patch embedding & $S \times D$ & Teacher \\
    $\mathbf{q}^{t}_{n}$ & Query of the $n$-th proto-object & $\mathbb{R}^{D_q}$ & Teacher \\
    $\mathbf{k}^{t}_{n}$ & Key of the $n$-th proto-object & $\mathbb{R}^{D_k}$ & Teacher \\
    $\mathbf{A}^{t}_{n}$ & Attention map of proto-object $n$ & $h \times w$ & Teacher \\
    \midrule
    \multicolumn{4}{l}{\textit{Method-specific Variables}} \\
    $w^{t}_{n}$ & Importance weight for proto-object $n$ & scalar & Teacher \\
    $\mathrm{Mask}^{t}_{n}$ & Spatial mask for proto-object $n$ & $H \times W$ & Teacher \\
    $M^{(t,t')}_{n}$ & Temporal validity mask & --- & Teacher \\
    $\mathcal{T}$ & Sampled proto-object subset & --- & --- \\
    $\mathcal{P}$ & Set of valid proto-objects ($\mathcal{L}_{\text{temp}}$) & --- & --- \\
    $T_{clip}$ & Total number of frames in the clip & integer & --- \\
    \bottomrule
\end{tabular}
\end{table*}

\subsection{Details of Class-Agnostic Proto-object Extraction}
\label{appendix:proto_extraction}

This section provides the formal vector-matrix formulations and dimensional specifications corresponding to the delineation process in Section \ref{sec:proto-object_delineation}.
Let $S$ denote the number of spatial tokens and $D$ the feature dimension.

We articulate the process in three formal steps:

\paragraph{1. Prototype Synthesis.}
For the $n$-th attention head, we synthesize a prototype vector $\mathbf{o}^t_n \in \mathbb{R}^{D}$ by aggregating the query tokens $\mathbf{q}^t_n \in \mathbb{R}^{S \times D}$ weighted by the spatial attention map $\mathbf{A}^t_n \in \mathbb{R}^{1 \times S}$.
To facilitate matrix operations, we formulate this as:
{
\begin{equation}
\mathbf{o}^t_n = (\mathbf{A}^t_n \cdot \mathbf{q}^t_n)^\top
\end{equation}
}
This operation collapses the spatial dimension $S$, distilling the head's attentional focus into a single global descriptor.

\paragraph{2. Soft Assignment via Normalized Similarity.}
We compute the alignment between the teacher's patch embeddings $\mathbf{e}^t \in \mathbb{R}^{S \times D}$ and the synthesized prototype. 
To robustly measure visual correspondence, we project the \textbf{normalized} patch embeddings onto the prototype vector:
{
\begin{equation}
\mathbf{M}^t_n = \frac{\mathbf{e}^t}{\|\mathbf{e}^t\|_2} \cdot \mathbf{o}^t_n \quad \in \mathbb{R}^{S}
\end{equation}
}
This formulation effectively captures the angular alignment between patch features and the proto-object concept, matching the logic of cosine similarity while preserving the magnitude information of the prototype.

\paragraph{3. Mask Generation \& Disentangled Feature Extraction.}
A binary mask is derived via parameter-free mean-thresholding: $\text{Mask}^t_n = \mathbb{1}(\mathbf{M}^t_n > \mathbb{E}[\mathbf{M}^t_n])$.
To ensure maximal feature disentanglement, this mask is \textbf{upsampled} to the image resolution and applied to the raw input pixels $X^t$. 
Let $\text{Up}(\cdot)$ denote the nearest-neighbor interpolation operator mapping spatial tokens to pixel coordinates:
{
\begin{equation}
\mathbf{f}^t_n = g_\theta(X^t \odot \text{Up}(\text{Mask}^t_n))
\label{eq:FinalRepresentation}
\end{equation}
}

\paragraph{Design Rationale: Why Pixel-level Masking?}
A critical methodological choice in Eq. \ref{eq:FinalRepresentation} is to apply the mask to the raw input $X^t$ rather than intermediate feature maps.
While feature-level masking is computationally cheaper, we prioritize \textbf{signal independence}.
In any shared backbone, the features of a specific patch inevitably aggregate information from the background and other objects due to expanding receptive fields or global self-attention mixing.
By forcing the student encoder to process the masked image from scratch, we physically block this ``information leakage.''
Although computationally sub-optimal, this design provides cleaner possible testbed for validating our core hypothesis regarding compositional consistency.

\paragraph{Rationale for Soft Assignment}
It is instructive to contrast our patch assignment logic with methods like DoRA~\cite{venkataramanan2023imagenet}.
DoRA utilizes the Sinkhorn-Knopp algorithm to enforce a competitive, ``winner-take-all" partition, meaning each patch must belong to exactly one slot. 
This structural rigidity can be limiting in cluttered egocentric scenes where occlusion and ambiguity are prevalent. 

In contrast, our approach employs a similarity-based soft assignment followed by independent thresholding.
We do not force prototypes to compete; a patch can be claimed by multiple heads or none at all.
This flexibility is better suited for unconstrained ``in-the-wild" video data, which necessitates a more permissive proto-object delineation strategy.

\subsection{Implementation Details of Proto-object Representation Learning}
\label{appendix:proto_loss_details}
This appendix provides the details for computing the compositional feature $\mathbf{f}^t_{\text{agg}}$, which is used in the second term $H(\mathbf{f'}, \mathbf{f}_{\text{agg}})$ of the $\mathcal{L}_{\text{proto}}$ objective (Eq.~\ref{eq:l_proto}). 

At each training iteration, for every video sequence $\{X^{t}\}_{t=1}^T$, we randomly sample a subset of attention heads $\mathcal{T} \subset \{1, \dots, N\}$ of size $K$ (set to $K=3$ in our experiments). 
This sampling strategy acts as a regularization mechanism, by forcing the student to form consistent scene representations from varying partial subsets of proto-objects.
\paragraph{Subset-wise Importance Weighting.}
For each head $n$ within the sampled subset $\mathcal{T}$, we compute its relative importance using the teacher's attention maps.
We first derive a saliency score $s^t_n$ by averaging the attention values over all spatial tokens (excluding the \texttt{[CLS]} token):
\begin{equation}
    s^t_n = \mathbb{E}_{\text{spatial}}[\mathbf{A}^t_n]
    = \frac{1}{S} \sum_{i=1}^{S} \mathbf{A}^t_{n,i}, \quad \forall n \in \mathcal{T},
\end{equation}
where $\mathbf{A}^t_n \in \mathbb{R}^{1 \times S}$ denotes the teacher's spatial attention map for head $n$ at time $t$, where $S$ is the number of spatial tokens.
Heads whose attention mass collapses to the \texttt{[CLS]} token tend to yield low spatial averages $s^t_n$, making $s^t_n$ a simple proxy for their semantic usefulness.
We then compute the normalized importance weights $w^t_n$ by applying a temperature-controlled softmax strictly over the sampled heads in $\mathcal{T}$:
\begin{equation}
    w^t_n = \frac{\exp(s^t_n / \tau_w)}{\sum_{k \in \mathcal{T}} \exp(s^t_k / \tau_w)}, \quad \forall n \in \mathcal{T},
\end{equation}
where $\tau_w$ is a temperature parameter (set to $0.1$ in our implementation), ensuring that the weights sum to $1$ within the current sampling context.

\paragraph{Local-to-Global Aggregation.}
Given the student's proto-object features $\{\mathbf{f}^t_n\}_{n \in \mathcal{T}}$, where each $\mathbf{f}^t_n \in \mathbb{R}^D$ is the head-specific proto-object representation defined in Eq.~\eqref{eq:FinalRepresentation}, we aggregate them using the derived weights:
\begin{equation}
    \mathbf{f}^t_{\text{agg}} = \sum_{n \in \mathcal{T}} w^t_{n} \cdot \mathbf{f}^t_{n}.
\end{equation}
The resulting vector $\mathbf{f}^t_{\text{agg}} \in \mathbb{R}^D$ serves as the student's reconstruction of the scene based on the selected proto-objects.

\subsection{Depth Loss Details}
\label{appendix:depth_loss_details}

To effectively leverage geometric priors, we employ a composite loss, $\mathcal{L}_{\text{depth}}$.
This loss is designed to distill structural information into our video encoder while remaining robust to the scale and shift ambiguities inherent in pseudo-labels.
It consists of a global alignment term ($\mathcal{L}_{\text{si}}$) and a gradient consistency term ($\mathcal{L}_{\text{grad}}$), both computed in the linear depth space consistent with our normalized output range $[0, 1]$.

The first component is the \textbf{Scale-Invariant Loss ($\mathcal{L}_{\text{si}}$)}.
We implement this term in linear space to specifically penalize relative structural errors while ignoring global offset discrepancies.
Let $d^t = \hat{D}^t - D^t$ be the pixel-wise difference between the predicted depth $\hat{D}^t$ and the teacher's depth prior $D^t$.
The loss is defined as:
\begin{equation}
\mathcal{L}_{\text{si}} = \frac{1}{\Omega}\sum_{i=1}^{\Omega}(d^t_i)^2 - \beta \cdot \left(\frac{1}{\Omega}\sum_{i=1}^{\Omega}d^t_i\right)^2
\end{equation}
Here, $\Omega$ denotes the total number of valid pixels in the frame $X^t$. % 明确定义 Omega
Following our implementation, we set $\beta = 0.5$.

The second component, the \textbf{Gradient Consistency Loss ($\mathcal{L}_{\text{grad}}$)}, encourages the student to capture high-frequency geometric details such as object boundaries.
We compute the Mean Absolute Error (MAE) between the spatial gradients of the prediction and the prior:
\begin{equation}
\begin{split}
\mathcal{L}_{\text{grad}}
&= \text{MAE}(\nabla_x \hat{D}^t, \nabla_x D^t) + \text{MAE}(\nabla_y \hat{D}^t, \nabla_y D^t) \\
&= \frac{1}{n_x}\sum_{i=1}^{n_x}\big|\nabla_x \hat{D}^t_i - \nabla_x D^t_i\big| \\
&\quad + \frac{1}{n_y}\sum_{i=1}^{n_y}\big|\nabla_y \hat{D}^t_i - \nabla_y D^t_i\big|
\end{split}
\end{equation}
where $\nabla_x$ and $\nabla_y$ are first-order finite difference operators.
$n_x$ and $n_y$ represent the number of valid gradient components in the horizontal and vertical directions, respectively. 

The total depth loss is the weighted sum $\mathcal{L}_{\text{depth}} = \mathcal{L}_{\text{si}} + \lambda_{\text{grad}}\mathcal{L}_{\text{grad}}$, with $\lambda_{\text{grad}}=1.0$.
This regularization explicitly guides the model to learn geometrically grounded representations without requiring manual annotations.

\subsection{Temporal Proto-object Consistency Learning}
\label{appendix:temporal_loss_details}
For clarity, we denote the second frame as $t'$, which corresponds to a temporal step $t+w$ relative to the first frame $t$, where $|w| \le W$ is the offset within the temporal window $W$.

To ensure representations are consistent over time, we introduce a teacher-filtered temporal contrastive loss. Within a temporal window $W$, for any two frames $t$ and $t'$, we first assess the reliability of correspondence using the teacher network. Specifically, we compute the cosine similarity between penultimate-layer features $\mathbf{z}'^t_n$ and $\mathbf{z}'^{t'}_n$:
\begin{equation}
\text{sim}(\mathbf{z}'^t_n, \mathbf{z}'^{t'}_n) = \frac{\mathbf{z}'^t_n \cdot \mathbf{z}'^{t'}_n}{\|\mathbf{z}'^t_n\|_2 \cdot \|\mathbf{z}'^{t'}_n\|_2}.
\end{equation}
We then define a validity mask $M^{(t,t')}_n$ by thresholding this similarity with a confidence threshold $\lambda$:
\begin{equation}
M^{(t,t')}_n = \mathbb{1}[\text{sim}(\mathbf{z}'^t_n, \mathbf{z}'^{t'}_n) > \lambda],
\end{equation}
where $\mathbb{1}[\cdot]$ is the indicator function. This mask is 1 only if the teacher deems the correspondence reliable.

For every valid pair $M^{(t,t')}_n=1$, we apply a contrastive loss.
We treat the student feature $\mathbf{z}^{t'}_n$ as the query, the matching teacher feature $\mathbf{z}'^{t}_n$ as the positive key, and all other teacher proto-object features at time $t$, $\mathbf{z}'^{t}_{k}$ for $k \neq n$, as negative keys.
The loss for this valid pair $(t, t')$ is:
{\small
\begin{equation}
\mathcal{L}^{(t, t')}_{\text{temp}} = -\frac{1}{|\mathcal{P}|} \sum_{n \in \mathcal{P}} \log \frac{
\exp(\text{sim}(\mathbf{z}^{t'}_n, \mathbf{z}'^t_n) / \alpha)}
{\sum_{k=1}^{\mathcal{K}} \exp(\text{sim}(\mathbf{z}^{t'}_n, \mathbf{z}'^t_k) / \alpha)}
\label{eq:Ltemp_appendix_unified}
\end{equation}
}
Here, $\mathcal{P} = \{n \mid M^{(t, t')}_n = 1\}$ is the set of valid proto-objects, $|\mathcal{P}|$ is the count of valid proto-objects, $\mathcal{K}$ is the total number of proto-objects, and $\alpha$ is a temperature parameter.

\paragraph{Total Temporal Loss Aggregation}
The total temporal loss $\mathcal{L}_{\text{temp}}$ is the average of the single-pair losses $\mathcal{L}^{(t, t')}_{\text{temp}}$ over all possible pairs within the clip. Assuming the clip length is $T_{clip}$, the final loss is defined as:
\begin{equation}
\mathcal{L}_{\text{temp}} = \frac{1}{N_{\text{pairs}}} \sum_{t=1}^{T_{clip}} \sum_{w=1}^{W} \mathcal{L}^{(t, t+w)}_{\text{temp}}
\end{equation}
where $N_{\text{pairs}}$ is the total number of valid sampled temporal pairs $(t, t+w)$ in the clip, constrained by $t+w \le T_{clip}$. Note that the summation uses the explicit time offset $w$ for clarity in defining the aggregation window.

\subsection{Complete Algorithm}
We present the complete algorithm of EgoViT in Algorithm \ref{alg:train-full}.
\begin{algorithm*}[t]
\caption{EgoViT -- Full Pseudocode of One Training Iteration}
\label{alg:train-full}
\small
\begin{algorithmic}[1]
    \State \textbf{Input:} Input video clip $\{X^t\}_{t=1}^{T_{clip}}$, student parameters $\theta$, teacher parameters $\theta'$.
    \State \textbf{Output:} Updated parameters $\theta$ and $\theta'$.

    \Statex \textit{1. Forward pass on original images for global features \& mask generation}
    \State \hspace{1.5em}\textbf{for} each frame $t$ in the clip $\{X^t\}_{t=1}^{T_{clip}}$ \textbf{do}
        \State \hspace{3em}$(\mathbf{f}^t, \mathbf{m}^t) \leftarrow \text{Student}_{\theta}(X^t)$ \Comment{Get final $\mathbf{f}^t$, and middle $\mathbf{m}^t$ features}
        \State \hspace{3em}$(\mathbf{f}^{\prime t}, \mathbf{A}^t) \leftarrow \text{Teacher}_{\theta'}(X^t)$ \Comment{Get features and attention maps}
    \State \hspace{1.5em}\textbf{end for}
    
    \Statex \textit{2. Generate a batch of masked images for proto-objects}
    \State \hspace{1.5em}Generate masks $\{\mathrm{Mask}^t_n\}$ from teacher attention maps $\{\mathbf{A}^t\}$.
    \State \hspace{1.5em}Create a set of masked images $\{P^t_n \leftarrow X^t \odot \mathrm{Mask}^t_n\}$.

    \Statex \textit{3. Forward pass on masked images to get object-specific features}
    \State \hspace{1.5em}$\{\mathbf{f}^t_n\}, \{\mathbf{z}^t_n\} \leftarrow \text{Student}_{\theta}(\{P^t_n\})$ \Comment{Individual proto-object final features $\mathbf{f}^t_n$ \& bottleneck features $\mathbf{z}^t_n$}
    \State \hspace{1.5em}$\{\mathbf{z}^{\prime t}_n\} \leftarrow \text{Teacher}_{\theta'}(\{P^t_n\})$ \Comment{Teacher bottleneck features for temporal filter $\mathbf{z}^{\prime t}_n$}

    \Statex \textit{4. Compute all loss terms}
    \State \hspace{1.5em}Compute depth loss $\mathcal{L}_{\text{depth}}$ from intermediate features $\{\mathbf{m}^t\}$.
    \State \hspace{1.5em}Compute proto-object loss $\mathcal{L}_{\text{proto}}$:
    \State \hspace{3em}a. Global alignment $H(\mathbf{f}^{\prime t}, \mathbf{f}^t)$.
      \State \hspace{3em}b. Compositional alignment $H(\mathbf{f}^{\prime t}, \mathbf{f}^t_{\text{agg}})$ using $\{\mathbf{f}^t_n\}$.
    \State \hspace{3em}Set $\mathcal{L}_{\text{proto}} = H(\mathbf{f}^{\prime t}, \mathbf{f}^t) + H(\mathbf{f}^{\prime t}, \mathbf{f}^t_{\text{agg}})$. 
    \State \hspace{1.5em}Compute temporal loss $\mathcal{L}_{\text{temp}}$ using filtered pairs from $\{\mathbf{z}^t_n\}$ and $\{\mathbf{z}^{\prime t}_n\}$.

    \Statex \textit{5. Joint Optimization and Model Update}
    \State \hspace{1.5em}Aggregate the total loss: $\mathcal{L}_{\text{total}} = \gamma_{P} \mathcal{L}_{\text{proto}} + \gamma_{D} \mathcal{L}_{\text{depth}} + \gamma_{T} \mathcal{L}_{\text{temp}}$. 
    \State \hspace{1.5em}Update student parameters $\theta$ via backpropagation on $\mathcal{L}_{\text{total}}$.
    \State \hspace{1.5em}Update teacher parameters $\theta'$ using an Exponential Moving Average (EMA) of $\theta$.
\end{algorithmic}
\end{algorithm*}

\section{Depth Decoder Architecture}
\label{appendix:depth_decoder}
In this section, we present a detailed description of the \textbf{Depth Decoder} architecture employed in our \textit{Depth Regularization} module.
The decoder is specifically designed to process intermediate feature representations $\mathbf{m}$ from the student encoder while maintaining computational efficiency. The input feature $\mathbf{m}$ has a shape of $C_{\text{in}} \times H' \times W'$ (where $C_{\text{in}} = D_1$ from Table \ref{tab:notation}).
Our decoder employs a progressive upsampling strategy, gradually recovering spatial details through multiple stages while reducing channel dimensionality.

\paragraph{Architecture Overview.}
The decoder begins with the input feature tensor $\mathbf{m}$ and processes it through an initial stage of refinement.
The architecture can be formally described as:
\begin{equation}
F_1 = \text{GELU}(\text{GN}_8(\text{Conv}_{3 \times 3}(\mathbf{m}, C_{\text{in}} \rightarrow 256)))
\end{equation}
where $\text{Conv}_{3 \times 3}(\text{C}_{\text{in}} \rightarrow 256)$ denotes a 2D convolution with kernel size 3 and padding 1, $\text{GN}_8$ represents Group Normalization with 8 groups, and GELU is the activation function.

\paragraph{Progressive Upsampling.}
The decoder comprises four sequential upsampling stages, each following the structure:
\begin{equation}
F_{i+1} = \text{GELU}(\text{GN}_{k_i}(\text{ConvTranspose}_{3 \times 3}(F_i)))
\end{equation}
$\text{for } i \in \{1, 2, 3, 4\}$, where $k_i = \min(8, C_{\text{out}})$ is the number of groups in Group Normalization, adaptively set based on the output channel dimension $C_{\text{out}}$.
The $\text{ConvTranspose}_{3 \times 3}$ operation uses a stride of 2, padding of 1, and an output padding of 1 to ensure the spatial resolution is exactly doubled at each stage.
The channel dimensions progressively decrease through the stages as follows: $256 \rightarrow 128 \rightarrow 64 \rightarrow 32 \rightarrow 16$.
\paragraph{Final Depth Prediction.}
The final depth map $\hat{D}^t$ is produced through a convolution with output channel 1, followed by a sigmoid activation to normalize the output to the range $[0, 1]$:
\begin{equation}
\hat{D}^t = \sigma(\text{Conv}_{3 \times 3}(F_5, 1))
\end{equation}
where $\sigma$ denotes the sigmoid activation function, and the final $\text{Conv}_{3 \times 3}$ outputs 1 channel.

\paragraph{Initialization Strategy.}
We employ He initialization for all convolutional and transposed convolutional layers to ensure stable training:
\begin{equation}
w \sim \mathcal{N}\left(0, \frac{2}{n_{\text{in}}}\right)
\end{equation}
where $n_{\text{in}}$ is the number of input units in the weight tensor. The Kaiming (He) initialization uses the assumption of a $\text{ReLU}$ non-linearity ($\text{nonlinearity=`relu'}$) as standard practice in PyTorch for deep convolutional networks.

\paragraph{Normalization and Activation.}
The use of Group Normalization instead of Batch Normalization makes the model more robust to varying batch sizes and provides consistent performance across different training configurations.
The GELU activation was chosen for its smooth characteristics and compatibility with transformer-based architectures.

\section{Experimental Details}
\label{appendix:implementation}
This appendix provides supplementary details for our experimental setup, including pre-training data, baselines, implementation hyperparameters, and downstream task protocols.

\subsection{Pre-training Datasets}
\label{sec:appendix_datasets}
We use videos from the Walking Tours (WT) dataset~\cite{venkataramanan2023imagenet}, which consists of 10 long-form, first-person videos captured in various cities worldwide. All videos were recorded at 4K resolution and 60 FPS, providing high-quality and temporally dense data for self-supervised pre-training. For our experiments, we selected a subset of these videos, as detailed in Table~\ref{tab:appendix_datasets}. This selection was made to expose our model to diverse visual environments, approximating open-world learning conditions.

\begin{table}[h]
\centering
\scriptsize
\caption{Details of egocentric videos used for pre-training. All videos are 4K at 60 FPS. The primary video is marked in bold; all others are used for augmentation.}
\label{tab:appendix_datasets}
\begin{tabular}{@{}lc@{}} 
\toprule
\textbf{Video Name} & \textbf{Duration (min)} \\ 
\midrule
\textbf{WT-Zurich} & $\mathbf{\sim 65}$ \\ 
WT-Istanbul & $\sim$73 \\
WT-Stockholm & $\sim$68 \\
WT-Chiang Mai & $\sim$70 \\
WT-Kuala Lumpur & $\sim$66 \\
WT-Venice & $\sim$110 \\
WT-Amsterdam & $\sim$82 \\
WT-Bangkok & $\sim$175 \\
WT-Singapore & $\sim$97 \\
WT-Wildlife & $\sim$60 \\
\bottomrule
\end{tabular}
\end{table}

\subsection{Architecture and Training Configuration}
\label{subsec:arch_config}
We adopt ViT-S/16~\citep{dosovitskiy2020image} as our backbone architecture, consisting of a 12-layer transformer with embedding dimension 384 and 6 attention heads per layer.
We randomly sample video clips of $T = 8$ frames with temporal separation of 1 second (i.e., one frame every 60 frames at 60 FPS) for each mini-batch.

To fully leverage high-resolution 4K video content, we first crop a 640×640 region from each frame at a scale from 0.4 to 1, and then apply a multi-crop strategy at a relatively small scale ranging from 0.15 to 0.3.
This scale range maintains the balance between object diversity and visual clarity. 
We use only two global crops per frame without additional local crops combined with cross-entropy loss, as videos inherently contain numerous irrelevant objects that could introduce excessive noise during training.
Additionally, we apply masking to the global crops fed into the student model to better facilitate learning of robust proto-object representations.

For optimization, we employ AdamW with base learning rate $\eta = 5 \times 10^{-4}$, initial weight decay $\lambda_{\text{wd}} = 0.04$, and linear warm-up over the first 10 epochs.
Given the temporal nature of video data, we define one epoch as a complete traversal of the video dataset and train for 320 epochs by default.
All experiments use a global batch size of 256.

\subsection{Proto-object Learning Hyperparameters}
\label{subsec:proto_hyperparams}

\begin{table}[h]
\centering
\scriptsize
\caption{Hyperparameters for proto-object learning framework.}
\label{tab:proto_hyperparams}
\begin{tabular}{lcc}
\toprule
\textbf{Parameter} & \textbf{Symbol} & \textbf{Value} \\
\midrule
Student Temperature & $\alpha_{\text{student}}$ & 0.1 \\
Teacher Temperature (Final) & $\alpha_{\text{teacher}}$ & 0.04 \\
Teacher Similarity Threshold & $\lambda$ & 0.8 \\
Sampled Proto-objects ($K$) & $K$ & 3 \\
Temporal Window Size & $W$ & 4 \\
Teacher momentum coefficient & $m$ & 0.996 \\
\bottomrule
\end{tabular}
\end{table}
Table~\ref{tab:proto_hyperparams} summarizes the key hyperparameters for our proto-object learning framework. Note that the Teacher Momentum $m$ follows a cosine schedule, increasing towards 1.0 during training.

\subsection{Downstream Task Evaluation Protocols}

\paragraph{Classification} 
We evaluate the classification capabilities of EgoViT and several baseline detection models pretrained on the Zurich dataset using ImageNet-1k as the downstream benchmark.
Specifically, EgoViT is pretrained on Zurich following the approach described in Sec. \ref{appendix:implementation}.
For fair comparisons, the baseline models are pretrained similarly, with an initial crop of 640×640 pixels taken from each frame, while subsequent data augmentations and training procedures follow the original baseline methodologies.
In the pretraining stage, we employ AdamW optimization with a global batch size of 256, a base learning rate of $5\times10^{-4}$, and a minimum learning rate of $1\times10^{-6}$.

For downstream classification evaluation, we perform two standard tasks: linear probing and k-nearest neighbor ($k$-NN) classification. 
In the linear probing setting, we follow the evaluation protocol of \citet{caron2021emerging}.
Specifically, we freeze the pretrained backbone features and train a linear classifier under supervised conditions on the ImageNet-1K training set, using a batch size of 1024. 
Performance is reported as top-1 accuracy (\%) on the ImageNet-1K validation set.
For $k$-NN classification, we again freeze the pretrained backbone to extract features from the ImageNet-1k training set and apply a k-nearest neighbor classifier with k=20. 
We report top-5 accuracy (\%) as the primary evaluation metric for comparison.

\paragraph{Object Discovery} 
Following LOST \cite{simeoni2021localizing}, we extract and average the self-attention maps from the final layer of our pretrained ViT-S/16, retaining 80\% of the total attention mass. We evaluate object localization performance on the Pascal VOC 2012 \cite{pascal-voc-2012} dataset, consisting of 11,540 images, using the CorLoc metric. CorLoc measures the localization accuracy as the percentage of correctly predicted bounding boxes, where a prediction is considered correct if its intersection-over-union (IoU) with the ground truth bounding box is greater than or equal to 0.5.

\paragraph{Object Detection and Instance Segmentation}
Due to computational constraints, we evaluate EgoViT for object detection and instance segmentation on the Mini COCO dataset \cite{samet2020houghnet}, a category-balanced subset of MS COCO \cite{lin2014microsoft} that effectively reflects model performance on the complete dataset. 
Specifically, we use ViT-S/16 as our backbone network, following the approach described in iBOT \cite{zhou2021ibot}, and apply a multi-scale training strategy. 
During training, input images are randomly resized, with their shorter sides ranging between 480 and 800 pixels while ensuring the longer side does not exceed 1333 pixels.
The entire network is fine-tuned using a standard $1\times$ schedule (12 epochs in total), with an initial learning rate of $1\times10^{-4}$, weight decay of 0.05, and learning rate decay by a factor of 10 at epochs 9 and 11. 
Moreover, we explore different layer-wise learning rate decay values, specifically \{0.65, 0.75, 0.8, 0.9\}, where a decay value of 1.0 indicates no layer-wise decay.

To construct hierarchical feature representations, we adapt the standard ViT-FPN conversion used in DINO.
We extract features from layers 4, 6, 8, and 12 of the backbone, mapping them to standard FPN levels ($P_2, P_3, P_4, P_5$ strides). 
Concretely, we perform two successive deconvolutions on the features from layer 4 to reach the highest resolution, a single deconvolution on layer 6 features, identity mapping on features from layer 8, and max-pooling to downsample features from layer 12. 
This process converts the single-scale ViT output into a multi-scale FPN suitable for detection and segmentation tasks.

\paragraph{Semantic Segmentation} 
For the Semantic Segmentation task, we fine-tune the model on ADE20K \cite{zhou2017scene} using a UperNet segmentation head for 160K iterations. Our experimental settings closely follow the procedure introduced in BEiT \cite{bao2021beit}. Specifically, we employ the AdamW optimizer with an initial learning rate of $6\times10^{-5}$ and a weight decay of $1\times10^{-2}$. A linear warm-up schedule is applied during the first 1,500 iterations. The model is fine-tuned with a batch size of 4.

\paragraph{Video Object Segmentation} 
For evaluating the performance of EgoViT on the video object segmentation task, we utilize the DAVIS 2017 dataset \cite{pont20172017}.
Following the evaluation protocol described in DINO \cite{caron2021emerging}, segmentation is performed on video frames at 480p resolution, each containing between two and four distinct objects.
We report performance using mean region-based similarity ($J_m$) and mean contour-based accuracy ($F_m$) metrics.

\section{On the Reproducibility of the DoRA Baseline}
\label{appendix:reproducibility}

To establish a fair and rigorous comparison, we made a significant effort to reproduce the results of our primary video-based baseline, DoRA. This section details our reproduction process and findings.

Our process was based on the official source code (commit hash: \texttt{DoRA\_ICLR24}) and we meticulously followed the experimental settings described in their paper, as detailed in our implementation setup in Sec.~\ref{appendix:implementation}.

Despite these efforts, we observed a notable discrepancy between our reproduced results and those reported in the original paper. This gap suggests a high sensitivity to specific, unstated details of the training environment or data preprocessing pipeline. For instance, on ADE20k semantic segmentation, our implementation achieved 21.6 mIoU, compared to the 35.4 mIoU reported in the original work. 
Similarly, for ImageNet-1k linear probing, we obtained 29.6\% accuracy, whereas the original work reported 44.5\%.

We note that this reproducibility challenge is not unique to our experience. Similar difficulties have been reported by other researchers in public forums, such as the issues section of the official DoRA GitHub repository (e.g., Issue \#1, \#3, \#4, and \#5).

Therefore, to maintain a controlled and scientifically valid comparison, all baseline results presented in our main paper are generated from our own implementation within a unified execution environment.
This ensures that the performance gains of our proposed method, EgoViT, are evaluated against a consistently implemented and directly comparable baseline, providing a true measure of its advancements.

\section{Additional Experimental Results}
\label{appendix:additional-experimental-results}

\begin{table}[t]
\centering
\scriptsize
\caption{Feature extraction depth evaluation on $k$-NN and CORLOC metrics. Shallower layers (depth 3–4) better preserve spatial information while maintaining comparable semantic features.}
\label{tab:ablation_depth}
\begin{tabular}{lccc}
\toprule
METHOD & Depth & $k$-NN & CORLOC \\
\midrule
DINO & $\times$ & 33.7 & 28.6 \\
EgoViT-D & 3 & 34.5 & \textbf{40.6} \\
EgoViT-D & 4 & \textbf{35.0} & 39.8 \\
EgoViT-D & 5 & 34.3 & 35.2 \\
EgoViT-D & 6 & 34.9 & 37.2 \\
EgoViT-D & 8 & 34.8 & 39.0 \\
EgoViT-D & 12 & 35.2 & 26.7 \\
\bottomrule
\end{tabular}
\end{table}

\begin{table}[t]
\centering
\scriptsize
\caption{Ablation on temporal modeling with different thresholds ($\lambda$) and temporal window sizes ($W$). CORLOC benefits from stricter correspondence filtering.}
\label{tab:ablation_threshold}
\begin{tabular}{lccc} 
\toprule
\multicolumn{4}{c}{\textbf{Threshold and Window Size Effect}} \\
\midrule
$W$ & $\lambda$ & $k$-NN & CORLOC \\
\midrule
3 & 0.5 & 20.8 & 32.2 \\
3 & 0.7 & 22.6 & 36.2 \\
3 & 0.8 & 22.7 & 35.9 \\
3 & 0.9 & 22.1 & 37.5 \\
4 & 0.7 & 22.2 & 37.0 \\
4 & 0.8 & \textbf{22.9} & 37.9 \\
4 & 0.9 & 22.2 & \textbf{38.0} \\
5 & 0.7 & 22.0 & 35.2 \\
5 & 0.9 & 22.0 & 35.4 \\
\bottomrule
\end{tabular}
\end{table}

\subsection{Ablation Study Details} 
\label{appendix:Ablation-Study-Details} 

\paragraph{Similarity threshold and window size analysis} 
Table \ref{tab:ablation_threshold} reveals that higher similarity thresholds ($\lambda$: 0.8-0.9) generally yield better performance in our temporal proto-object learning module.
This suggests that stricter matching criteria lead to more reliable temporal associations between proto-objects, supporting our design choice to focus on high-confidence object correspondences across frames.
We also observe that classification performance ($k$-NN) is less sensitive to the specific threshold, achieving its optimum at $W=4$ and $\lambda=0.8$ ($22.9\%$).

\paragraph{Feature extraction depth} 
As shown in Table \ref{tab:ablation_depth}, the choice of feature extraction layer significantly impacts model performance.
Shallower layers (depth 3-4) yield optimal results for both metrics, substantially outperforming the DINO baseline.
Interestingly, the deepest layer (12) maintains strong classification performance but performs poorly on localization ($26.7\%$ $\text{CORLOC}$). 
This suggests a clear trade-off: shallower layers preserve spatial information critical for object localization, while deeper layers capture abstract semantic features beneficial for classification.

\paragraph{Qualitative Analysis of Learned Components} 
\begin{figure}
    \centering
    \caption{Ablation visualization of EgoViT.}
    \includegraphics[width=\columnwidth]
    {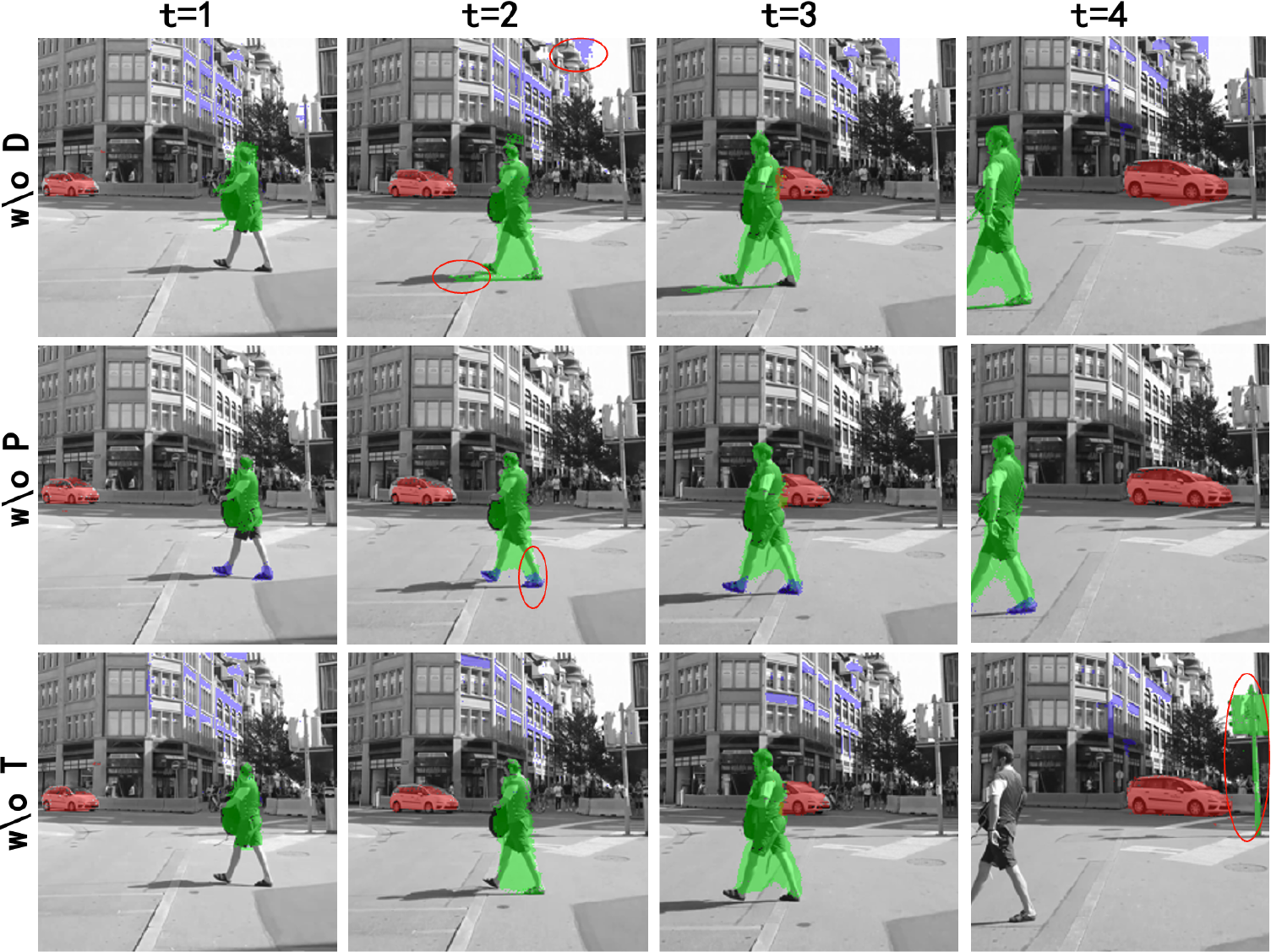} 
    \label{fig:visualization_2}
\end{figure}

Fig.~\ref{fig:visualization_2} shows: 
w/o D proto-objects attach to more background, w/o P heads attend to the same object, and w/o T assignments become unstable across frames.

\subsection{Other Experimental Results}
\label{appendix:Additional-Experimental-Results}
We present the comprehensive experimental results of EgoViT compared to state-of-the-art self-supervised methods. Our analysis is structured around quantifying the performance benefits derived from our \textbf{Proto-Consistency} paradigm, particularly focusing on object localization and temporal generalization.

\subsubsection{Performance on Downstream Tasks}

\begin{table*}[t]
\centering
\scriptsize
    \caption{
    Performance comparison of EgoViT against state-of-the-art methods across a range of downstream tasks. 
    Our main model, \textbf{EgoViT$_{\text{Zurich}}$}, is trained on a single 65-minute video, while EgoViT$_{\text{WT-Sub5}}$ is an additional model trained on five videos to demonstrate scalability.
    Models without an explicit subscript are trained on the default Zurich video unless otherwise noted. Except for PooDLe, which uses a ResNet-50 backbone (weights obtained from the official release), all other baselines and our models use a ViT-S architecture for fair comparison.
    }
    \vspace{-10pt}
    \begin{tabular}{l|cc|c|c|c|c|c|c|c|c}
    \hline
    \multirow{2}{*}{METHOD} & \multicolumn{2}{c|}{Semantic SEG.} & \multicolumn{1}{c|}{Object DET.} & \multicolumn{1}{c|}{Instance SEG.} & \multicolumn{3}{c|}{Video Object Segmentation} &  \multicolumn{1}{c|}{Object DIS.}& \multicolumn{2}{c}{Classification} \\
    \cline{2-11}
    & mIoU & Acc$_m$ & mAP  & mAP & $(\mathcal{J}\&\mathcal{F})_m$ & $\mathcal{J}_m$ & $\mathcal{F}_m$  & CORLOC & LP & $k$-NN \\
    \hline
    SimCLR & 22.5 & 32.8 & 22.2 & 20.1 & 52.3 & 51.5 & 53.1 & 35.5 & 28.2 & 33.1 \\
    AttMask  & 25.1 & 35.4 & 25.9 & 23.9 & 52.2 & 52.5 & 51.8 & 37.8 & 25.4 & 35.9 \\
    MoCo-v3  & 17.9 & 26.0 & 19.0 & 17.3 & 52.5 & 50.9 & 54.0 & 43.1 & 22.5 & 31.2  \\
    MAE  & 23.0 & 33.2 & 24.6 & 22.1 & 51.3 & 50.3 & 52.2 & 35.9 & 13.0 & 17.8  \\
    iBOT  & 23.9 & 34.5 & 22.1 & 19.6 & 53.9 & 53.5 & 54.3 & 36.3 & 27.5 & 33.6 \\
    SAVi++  & 24.4 & 35.7 & 24.7 & 23.4 & 52.0 & 50.8 & 53.2 & 34.2 & 29.3 & 31.2  \\
    DORA  & 21.6 & 31.1 & 22.6 & 20.4 & 53.8 & 51.9 & 55.6 & 24.1 & 29.6 & 33.7  \\
    DORA$_{\text{Venice}}$  & 22.4 & 32.3 & 23.2 & 21.2 & 53.0 & 51.7 & 54.2 & 23.9 & 30.2 & 34.8  \\
    PooDLe$_{\text{Venice}}$  & 33.5 & 41.6 & 26.2 & 23.7 & 19.7 & 21.4 & 17.8 & 32.7 & 28.9 & 30.8  \\
    \hline
    DINO  & 21.2 & 30.3 & 22.0 & 20.6 & 53.8 & 52.5 & 55.1 & 37.2 & 30.9 & 35.5  \\
    EgoViT$_{\text{Venice}}$ & 27.1 & 37.5 & 26.2 & 24.6 & 54.5 & 52.7 & 56.4 & 45.4 & 35.8 & 39.2 \\
    EgoViT & 26.0(\textcolor[rgb]{0.2,0.6,0.2}{+4.8}) & 36.6 (\textcolor[rgb]{0.2,0.6,0.2}{+6.3}) & 26.7 (\textcolor[rgb]{0.2,0.6,0.2}{+4.7}) & 24.3(\textcolor[rgb]{0.2,0.6,0.2}{+3.7}) & 54.3 (\textcolor[rgb]{0.2,0.6,0.2}{\scriptsize +0.5}) & 52.7 (\textcolor[rgb]{0.2,0.6,0.2}{\scriptsize +0.2}) & 55.9 (\textcolor[rgb]{0.2,0.6,0.2}{\scriptsize +0.8}) & 45.2 (\textcolor[rgb]{0.2,0.6,0.2}{\scriptsize +8.0}) & 34.0 (\textcolor[rgb]{0.2,0.6,0.2}{\scriptsize +3.1}) & 38.9 (\textcolor[rgb]{0.2,0.6,0.2}{\scriptsize +3.4}) \\
    EgoViT$_{\text{WT-Sub5}}$ & \textbf{29.0}(\textcolor[rgb]{0.2,0.6,0.2}{+7.8}) & \textbf{38.7} (\textcolor[rgb]{0.2,0.6,0.2}{+8.4}) & \textbf{28.2} (\textcolor[rgb]{0.2,0.6,0.2}{+6.2}) & \textbf{26.2}(\textcolor[rgb]{0.2,0.6,0.2}{+5.6}) & \textbf{56.1} (\textcolor[rgb]{0.2,0.6,0.2}{\scriptsize +2.3}) & \textbf{54.2} (\textcolor[rgb]{0.2,0.6,0.2}{\scriptsize +1.7}) & \textbf{57.8} (\textcolor[rgb]{0.2,0.6,0.2}{\scriptsize +2.7}) & \textbf{48.9} (\textcolor[rgb]{0.2,0.6,0.2}{\scriptsize +11.7}) & \textbf{37.2} (\textcolor[rgb]{0.2,0.6,0.2}{\scriptsize +6.3}) & \textbf{42.7} (\textcolor[rgb]{0.2,0.6,0.2}{\scriptsize +7.2})\\
    EgoViT$_{\text{WT-all}}$ & \textbf{30.6}(\textcolor[rgb]{0.2,0.6,0.2}{+9.4}) & \textbf{39.3} (\textcolor[rgb]{0.2,0.6,0.2}{+9.0}) & \textbf{29.6} (\textcolor[rgb]{0.2,0.6,0.2}{+7.6}) & \textbf{26.8}(\textcolor[rgb]{0.2,0.6,0.2}{+6.2}) & \textbf{57.0} (\textcolor[rgb]{0.2,0.6,0.2}{\scriptsize +3.2}) & \textbf{55.0} (\textcolor[rgb]{0.2,0.6,0.2}{\scriptsize +2.5}) & \textbf{58.9} (\textcolor[rgb]{0.2,0.6,0.2}{\scriptsize +3.8}) & \textbf{50.2} (\textcolor[rgb]{0.2,0.6,0.2}{\scriptsize +13.0}) & \textbf{39.1} (\textcolor[rgb]{0.2,0.6,0.2}{\scriptsize +8.2}) & \textbf{45.3} (\textcolor[rgb]{0.2,0.6,0.2}{\scriptsize +9.8})\\
    \hline
    \end{tabular}
    \label{tab:main_results_app}
\end{table*}

In this section, we provide additional comparisons that were excluded from the main text to ensure strictly controlled experimental conditions. We focus on two aspects: (1) comparison with specialized architectures, and (2) the scalability of EgoViT across different data regimes.

\paragraph{Comparison with Specialized Architectures (SAVi++, PooDLe).}
In the main paper, we restricted our comparison to methods utilizing standard ViT-S backbones.
Here, we extend the evaluation to include SAVi++ and PooDLe in the discussion. The former is a representative work that relies on slots for self-supervised learning, and the latter is a self-supervised paradigm aimed at object segmentation on ResNet-50.
As shown in Table \ref{tab:main_results_app}, although SAVi++ uses slots for object segmentation in its architecture, EgoViT$_{\text{Zurich}}$, based on the standard ViT, unexpectedly outperforms it in the segmentation task.
This result suggests that our Proto-Consistency objective effectively induces object-centric features within standard transformer architectures, without requiring complex slot-based modules. 
While PooDLe shows strong performance in Semantic Segmentation, it significantly underperforms in temporal tasks (VOS), whereas EgoViT maintains balanced performance across all metrics. 
Notably, the official PooDLe paper reports in its appendix that the ViT-S variant tends to collapse during video-based self-supervised training, which is why their semantic segmentation evaluation relies on a ResNet-50 backbone instead. 
Since ResNet-50 typically exhibits stronger performance than ViT-S under similar settings, this backbone discrepancy may partially account for PooDLe outperforming our EgoViT on semantic segmentation tasks.

\paragraph{Scalability Analysis (Sub-5 and Full Data).}
To demonstrate the data efficiency and scalability of our approach, we evaluate EgoViT on three progressively larger data scales:
\begin{enumerate}
    \item \textbf{Single Video:} Models trained on individual scenes (\textit{Zurich}, \textit{Venice}).
    \item \textbf{WT-Sub5 (5 Videos):} An intermediate scale trained on a curated subset of five videos to test generalization across diverse environments. This subset includes: \textit{Zurich, Venice, Istanbul, Stockholm, and Chiang Mai}.
    \item \textbf{WT-All (Full Dataset):} The model trained on the entire available dataset.
\end{enumerate}

\noindent \textbf{Trend Analysis:} The bottom section of Table \ref{tab:main_results_app} reveals a consistent upward trend. Moving from single-video training to the WT-Sub5 set yields immediate gains (e.g., $+3.0\%$ mIoU over Zurich), confirming that the model benefits from increased visual diversity. 
Furthermore, EgoViT$_{\text{WT-All}}$ achieves the best overall performance, indicating that our framework scales effectively with data volume and has not yet reached saturation.

\subsubsection{Temporal Generalization on Object Tracking}

\begin{table}[t]
\centering
\scriptsize
\caption{
DINO, DoRA, and EgoViT (Zurich-pretrained) backbones are evaluated under the OSTrack framework on TrackingNet and GOT-10k. 
}
\vspace{-6pt}
\begin{tabular}{lccc}
\toprule
\multicolumn{4}{c}{\textbf{TrackingNet}}\\
\cmidrule(lr){1-4}
Metric & DINO & DoRA & EgoViT \\
\midrule
AUC     & 76.2 & 77.7 & \textbf{78.9} \\
P\_Norm  & 81.2 & 82.5 & \textbf{83.5} \\
P       & 73.7 & 74.7 & \textbf{77.6} \\
\midrule
\multicolumn{4}{c}{\textbf{GOT-10k (zero-shot)}}\\
\cmidrule(lr){1-4}
Metric & DINO & DoRA & EgoViT \\
\midrule
AO       & 61.6 & 63.8 & \textbf{67.0} \\
SR$_{0.5}$ & 71.9 & 73.8 & \textbf{77.0} \\
SR$_{0.75}$& 59.5 & 62.4 & \textbf{65.5} \\
\bottomrule
\end{tabular}
\label{tab:generalization_trackingnet_got10k}
\end{table}

\paragraph{Validation through Object Tracking.}
Table \ref{tab:generalization_trackingnet_got10k} evaluates the robustness of the learned features on two standard tracking benchmarks, TrackingNet and GOT-10k, using the OSTrack framework. 
On TrackingNet, $\text{EgoViT}$ achieves the best performance across all three metrics, with an AUC of $78.9\%$ and consistently higher P and P\_Norm scores than both DINO and DoRA. 
On GOT-10k in the challenging zero-shot setting, the gap is even clearer: $\text{EgoViT}$ attains an AO of $67.0\%$ together with the highest SR$_{0.5}$ and SR$_{0.75}$, indicating stronger generalization to unseen targets and motion patterns.

These consistent improvements across two datasets with different object categories and motion patterns provide strong evidence that our design, which combines \emph{explicit temporal consistency supervision} with a depth-based geometric regularizer, leads to more robust and persistent object representations than purely spatial or view-based consistency. 
The resulting geometry-regularized and temporally filtered prototypes learned by $\text{EgoViT}$ transfer effectively to downstream trackers, enabling more reliable cross-frame identity association and long-term tracking.

\subsubsection{Performance on Egocentric Video Benchmark: Epic‑Kitchens VISOR}
We follow the Epic-Kitchens VISOR VOS protocol~\cite{darkhalil2022epic}, with only the backbone replaced. 
Table~\ref{tab:downstream_visor} shows EgoViT significantly outperforms baselines (e.g., +6.1\% in $\mathcal{J}\&\mathcal{F}$; +6.9\% on unseen subset).
These gains indicate that EgoViT learns stable representations for egocentric perception.

\begin{table}
    \centering
    \scriptsize
    \setlength{\tabcolsep}{3.5pt}
    \caption{EPIC-KITCHENS VISOR VOS results.
    All methods use a ViT-S/16 backbone.}
    \begin{tabular}{lcccccc}
        \hline
        Method & Pretrain & VOS Fine-Tune & $\mathcal{J}\&\mathcal{F}$ & $\mathcal{J}$ & $\mathcal{F}$ & $(\mathcal{J}\&\mathcal{F})_{unseen}$ \\
        \hline
        \textit{Baseline} & -- & $\checkmark$ & 55.5 & 53.9 & 57.1 & 49.5 \\
        DNIO  & WT$_{\text{Zurich}}$ & $\checkmark$ & 62.4 & 61.5 & 63.3 & 56.2 \\
        DORA  & WT$_{\text{Zurich}}$ & $\checkmark$ & 63.3 & 62.2 & 64.3 & 57.4 \\
        EgoViT  & WT$_{\text{Zurich}}$ & $\checkmark$ & \textbf{69.4} & \textbf{67.8} & \textbf{70.9} & \textbf{64.3} \\
        \hline
    \end{tabular}
    \label{tab:downstream_visor}
\end{table}

\section{More Visualization}
\begin{figure*}[htbp]
  \centering
  \includegraphics[width=\linewidth,page=1]{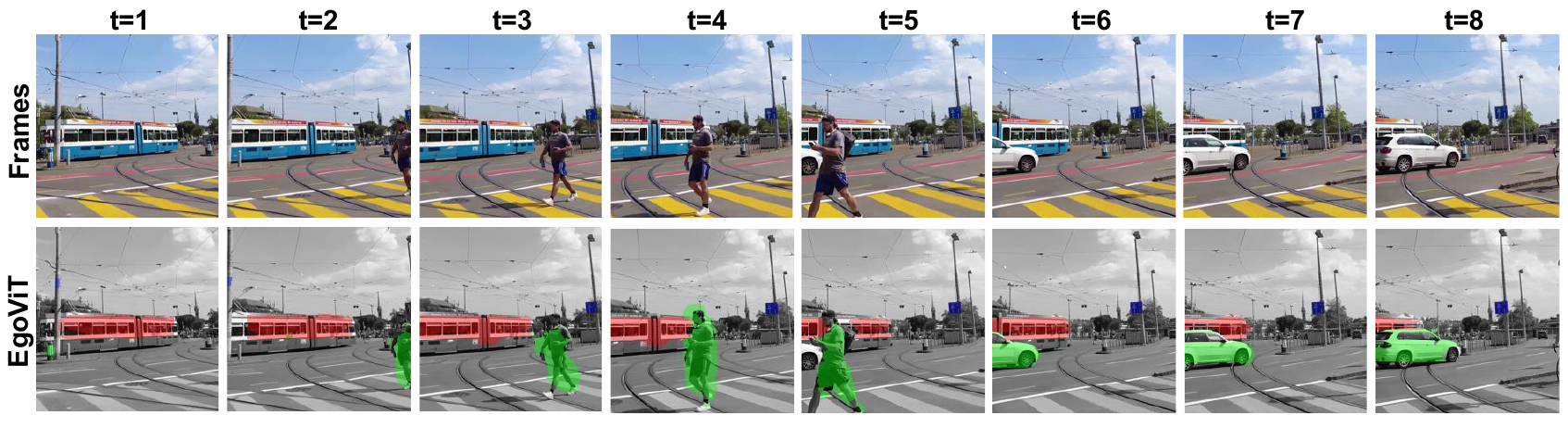}
  \caption{EgoViT maintains stable and distinct masks for the moving tram (red) and the pedestrian (green), effectively distinguishing them despite rapid camera panning and a cluttered background. When the pedestrian leaves the frame, the model seamlessly transitions to tracking the approaching car (also green). Additionally, EgoViT potentially exhibits partial capability in attending to small, distant objects, such as the signage (blue).}
  \label{fig:exp_page1}
\end{figure*}

\begin{figure*}[htbp]
  \centering
  \includegraphics[width=\linewidth,page=2]{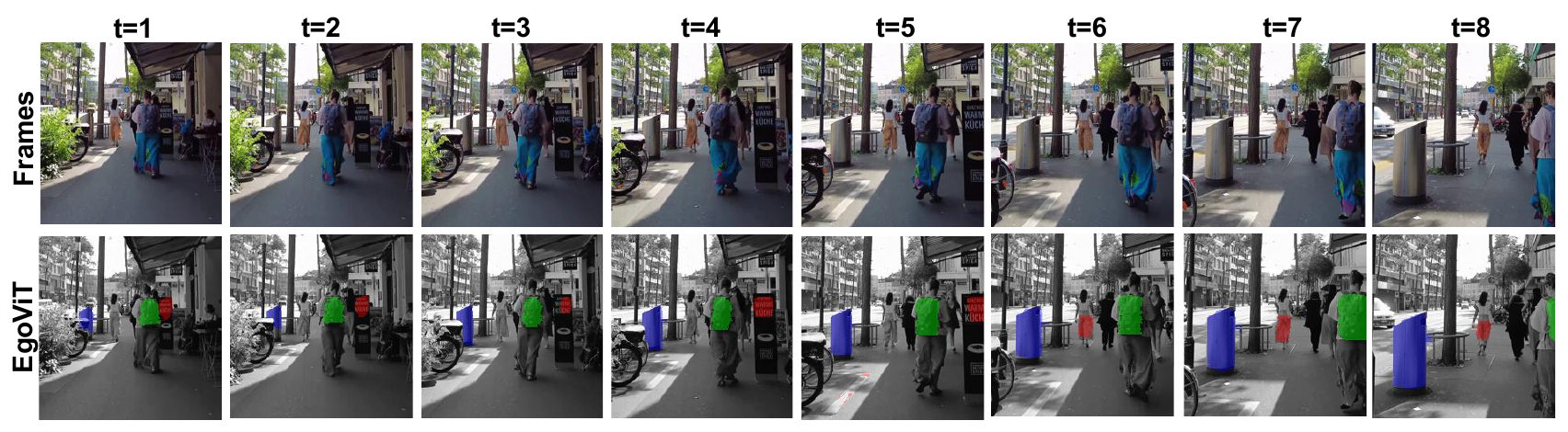}
    \caption{EgoViT maintains identity-consistent masks for the walking person (green) and the trash bin (blue), even as illumination, viewpoint, and crowd density vary. The text on the signage (red) is also consistently segmented across frames.}
  \label{fig:exp_page2}
\end{figure*}

\begin{figure*}[htbp]
  \centering
  \includegraphics[width=\linewidth,page=3]{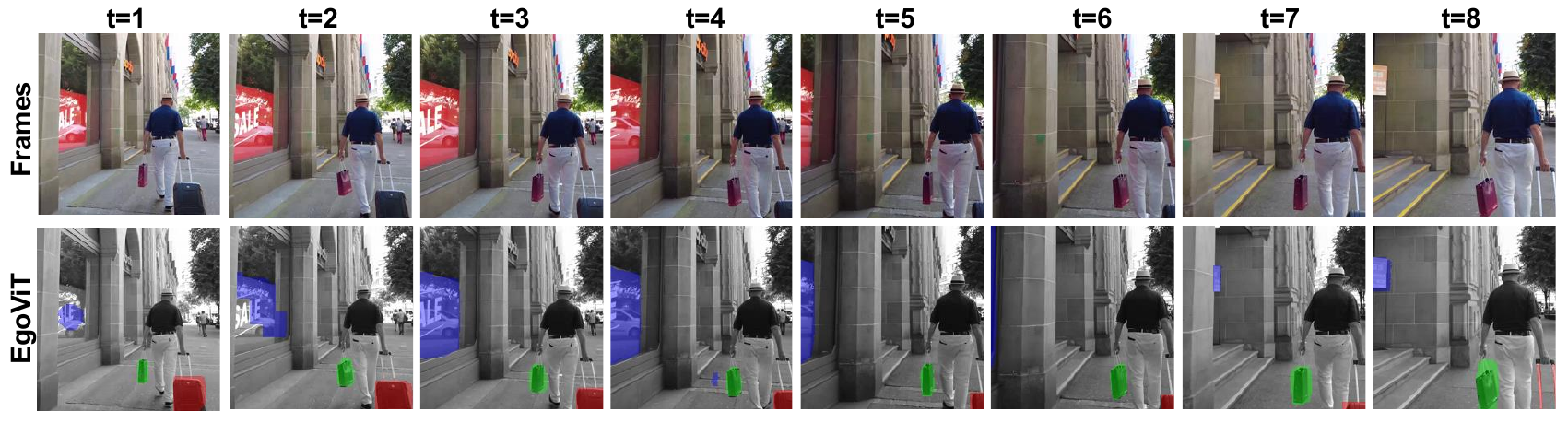}
  \caption{Even when the camera turns and the target becomes partially occluded, EgoViT tends to preserve relatively distinct proto-objects, such as the store sign (red), the shoulder bag (blue), and the suitcase (green). This may indicate that the model is robust to moderate viewpoint changes and occlusions.}
  \label{fig:exp_page3}
\end{figure*}

\paragraph{Qualitative analysis of temporally consistent proto-objects.}
Figure~\ref{fig:exp_page1}–\ref{fig:exp_page3} visualize EgoViT's predictions across 8 consecutive egocentric frames for diverse scenarios with dynamic objects and viewpoint shifts.

In Figure~\ref{fig:exp_page1}, EgoViT robustly maintains object identity for both a moving tram (red) and an approaching car (green), despite abrupt camera panning and strong background clutter (e.g., crosswalk stripes and shadows). The persistence of masks demonstrates the model's ability to filter motion-independent structure by leveraging depth and teacher-guided consistency.

In Figure~\ref{fig:exp_page2}, EgoViT successfully segments a person (green) and a luggage trolley (blue) in a crowded station scene with significant occlusion and illumination shifts. Importantly, the assigned masks remain identity-consistent even as both objects deform or partially disappear, showing that EgoViT encodes proto-objects beyond mere appearance.

Figure~\ref{fig:exp_page3} further highlights EgoViT's capacity to disambiguate multiple overlapping proto-objects (signboard, bag, suitcase), despite their similar texture and partial occlusions across frames. This illustrates the effectiveness of our depth-anchored proposal and temporal filtering modules in learning object-centric representations under self-supervision.

Overall, these results show that EgoViT goes beyond spatial saliency or appearance clustering: it learns \textbf{temporally persistent, semantically coherent object-level concepts} from egocentric video without requiring class labels.
\label{appendix:more_visualization}

\section{Broader Impacts and Ethical Considerations}
\label{appendix:broader-impacts}
Our work on EgoViT introduces a self-supervised framework for learning temporally consistent object representations from egocentric video streams. This capability has implications across embodied AI, cognitive modeling, and potential real-world deployments.
\textbf{We emphasize that EgoViT is presented as a research prototype designed to advance fundamental understanding in self-supervised learning.}

\subsection{Potential Positive Impacts}

EgoViT offers a foundation for more perceptually grounded embodied agents.
By learning to track spatially coherent entities over time without supervision, EgoViT enables downstream models to develop object permanence and identity persistence—capabilities critical for long-horizon interaction, manipulation, and navigation in real-world environments.

Additionally, our approach reduces reliance on large-scale manual annotations, making it suitable for deployment in novel, open-ended scenarios where semantic labels are scarce or costly to obtain (e.g., in-situ robotic learning or home-scale exploration). 
The model's biologically inspired structure—linking depth cues with temporal attention—also provides a computational tool that may inform studies in human perception.

\subsection{Risks and Mitigation Strategies}

Egocentric visual data inherently contains sensitive information about individuals and personal environments. 
Deploying systems like EgoViT without safeguards may lead to privacy breaches, especially through bystander re-identification or context inference.
We recommend future applications of EgoViT incorporate: (1) on-device processing, (2) anonymization pipelines (e.g., face blurring), and (3) user-controlled data access policies.

Further, the ability to stably track objects and infer scene structure could be misused in surveillance contexts.
While EgoViT is intended for research and interaction-based learning, we advise usage restrictions and transparent model cards to guide ethical downstream applications.

\subsection{Responsible Development}
To promote responsible use, we release our code and models with clear licensing terms that discourage surveillance use.
We are committed to continuing research in privacy-preserving self-supervised learning and encourage community engagement to identify and mitigate emergent risks.

\section{Detailed Comparison and Positioning Analysis with DINO and DoRA}
\label{app:detailed Comparison}
This appendix aims to precisely articulate the technical inheritance and paradigmatic distinction between our work and two pivotal prior works: DINO and DoRA. 
Our objective is to eliminate ambiguity regarding the contribution of this work.

\subsection{Foundational Framework: Build upon DINO} 
\label{sec:dino_inheritance} 
The foundational training framework of EgoViT is built upon the self-distillation mechanism proposed in DINO~\cite{caron2021emerging}.
We explicitly inherit its core Teacher-Student architecture, including the Exponential Moving Average (EMA) for parameter updates, the knowledge distillation loss, and the centering and sharpening strategies for the teacher's outputs.
We adopt this mature framework to ensure training stability and efficiency. 
Our core innovation lies in the novel supervisory signals we provide to this framework.

\subsection{Core differences between EgoViT and DoRA} 
\label{appendix:Core-differences-between-EgoViT-and-DoRA}
The most fundamental distinction between EgoViT and DoRA lies in the self-supervised learning paradigms they follow.
DoRA adopts a \textit{Multi-view Spatial Consistency} paradigm as its core learning principle, whereas EgoViT introduces a new paradigm we term \textit{Proto-Consistency}.

\paragraph{The DoRA Paradigm: Multi-view Spatial Consistency}
DoRA extends DINO's concepts from static images to video.
Its core paradigm can be summarized as \textbf{Multi-view Spatial Consistency}. 
The supervisory signal primarily originates from the alignment of different spatial views generated from the same point in time:
\begin{itemize}
    \item \textbf{Local-to-Global Alignment:} Requires features extracted from random local crops to align with features from a global view.
    \item \textbf{Masked-to-Global Alignment:} Utilizes self-attention maps to track salient image patches, generates a masked view, and requires features from this view to align with those of the global view.
\end{itemize}
Essentially, DoRA constructs its self-supervisory signal by creating and aligning different ``views", making inter-view consistency the core of its learning process.

\paragraph{The EgoViT Paradigm: Proto-Consistency}
We posit that in the complex scenarios of egocentric video, view-based alignment faces significant challenges. 
Therefore, EgoViT introduces a new paradigm of \textbf{Proto-Consistency}.
Our core objective is not to align different views, but to learn a robust set of prototypes and enforce their consistency across multiple dimensions, particularly over time.
This principle forms the cornerstone of our multi-task learning framework, manifesting as:
\begin{itemize}
    \item \textbf{Spatial Consistency:} Our `proto\_loss' requires the prototype features, obtained via soft aggregation, to be consistent with the global scene representation perceived by the teacher network.
    \item \textbf{Temporal Consistency:} Our `temporal\_loss' leverages cross-frame contrastive learning to ensure that the prototype representation of an object remains stable and consistent as time progresses.
\end{itemize}
In essence, the core of EgoViT's learning is the consistency of the prototype itself, which is jointly supervised across spatial and temporal dimensions through our multi-task objective.

\paragraph{Technical Implementation: Code-Level Evidence for the Two Paradigms}
This paradigmatic difference is substantiated by the causal relationship between \textbf{masked img} and \textbf{proto} in the respective implementations:
\begin{itemize}
    \item In DoRA, the \textbf{proto} is the \textbf{cause} (a template for mask generation), and the \textbf{masked img} is the \textbf{effect} (the final product for data augmentation). 
    The entire process serves to create a new ``view".
    \item In EgoViT, the \textbf{masked img} is the \textbf{cause} (a proposal to delineate foreground regions), and the \textbf{proto} is the \textbf{effect} (a direct learning objective that is optimized in the feature space). The entire process serves to learn the prototype itself.
\end{itemize}

\begin{table*}[t]
    \centering
    \caption{Conceptual comparison between DoRA and EgoViT.}
    \label{tab:dora_egovit_comparison}
    \small
    \setlength{\tabcolsep}{3pt}
    \begin{tabularx}{\textwidth}{lXX}
        \toprule
        \textbf{Dimension} & \textbf{DoRA} & \textbf{EgoViT (ours)} \\
        \midrule
        \textbf{Core paradigm} 
        & Multi-view spatial consistency on global image embeddings. 
        & Proto-consistency on object-centric prototypes distilled from attention heads. \\
        
        \textbf{Learning unit} 
        & Whole-scene representation; no explicit object-level carrier. 
        & Proto-object prototype as the basic unit for representation and supervision. \\
        
        \textbf{Temporal modeling} 
        & Similar patches across time used only to create augmented views; the loss is purely spatial, without explicit temporal consistency. 
        & Dedicated temporal loss that penalizes prototype drift and enforces identity persistence across frames. \\
        
        \textbf{Structure awareness} 
        & RGB-only, without geometric priors. 
        & Auxiliary depth regularization encourages geometry-aware, structure-sensitive proto-object representations. \\
        \bottomrule
    \end{tabularx}
\end{table*}

\paragraph{Conclusion}
In summary, the two methods differ fundamentally in their core paradigms (Multi-view Consistency vs. Proto-Consistency), learning objectives (single-task vs. multi-task), and underlying design philosophies, as summarized in Table~\ref{tab:dora_egovit_comparison}.

\section{Zero-Shot Generalization: Ego4D Case Study}
\label{appendix:ego4d}

\subsection{Motivation}
Our main pre-training setup uses a single long Zurich city-walk video. 
This deliberately minimalist training domain naturally raises the question of how well the learned representations generalize to more diverse egocentric environments.
In particular, we are interested in (i) larger egocentric corpora such as Ego4D, and (ii) scenarios with substantially different dynamics and visual conditions, such as cluttered indoor scenes and in-car low-light conditions. 
To investigate this, we conduct an additional zero-shot case study on Ego4D that primarily focuses on qualitative behavior: we keep the pre-training protocol fixed (single Zurich city-walk video) and only change the evaluation domain.

\subsection{Ego4D-mini Benchmark and Setup}

\begin{figure}[t]
  \centering
  \includegraphics[width=\linewidth]{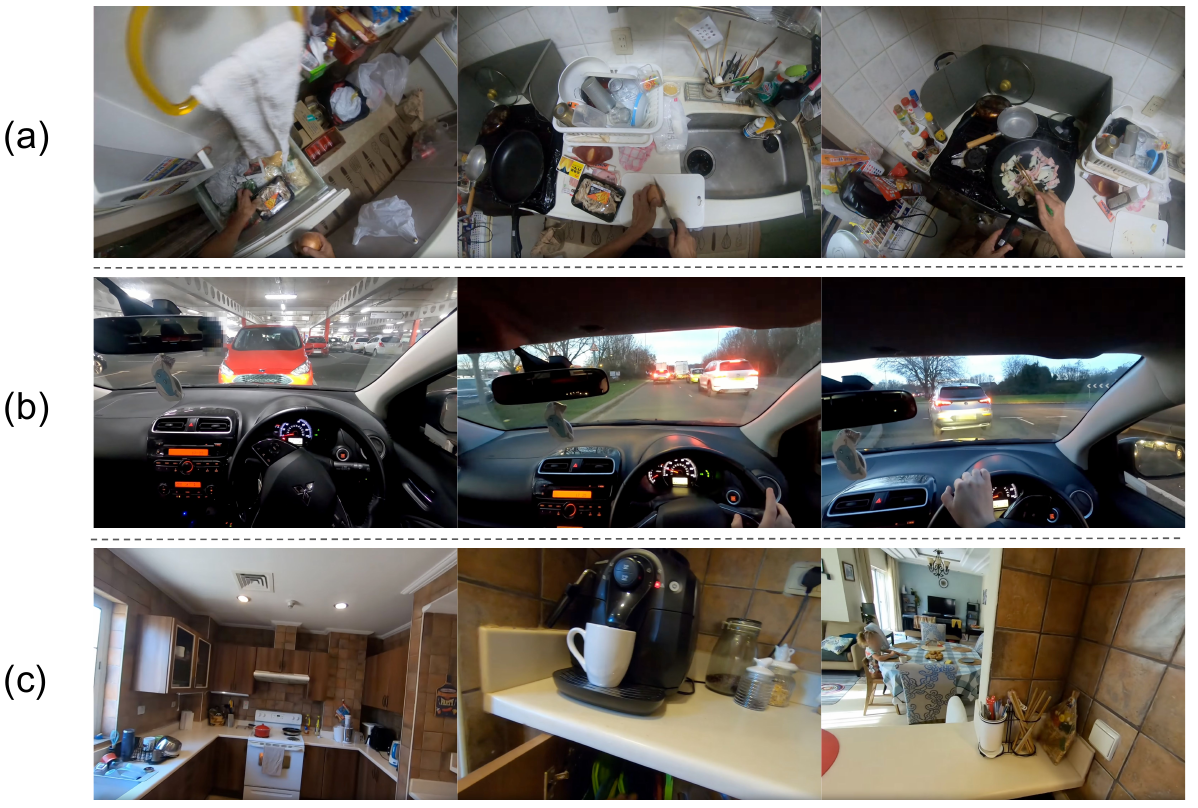}
    \caption{
    Example frames from the proposed Ego4D-mini benchmark.
    Row (a) shows cluttered kitchen scenes with strong hand--object interactions.
    Row (b) shows in-car driving sequences under low light and motion blur, with diverse dashboard and road objects.
    Row (c) shows domestic walking scenes.
    These clips are used only for evaluation; both DINO and EgoViT are pre-trained on a separate Zurich city-walk video.
    }
  \label{fig:ego4d_mini}
\end{figure}

We construct a small but deliberately diverse Ego4D-mini benchmark by selecting three long clips from Ego4D(example frames are shown in Fig.~\ref{fig:ego4d_mini}):
\begin{itemize}
\item \textbf{E4D-Kitchen}: indoor scenes with strong hand--object interaction and heavy background clutter(ID: 2422d726-0286-48bc-96a6-fe29c45cc409);
\item \textbf{E4D-Driving}: in-car sequences with low-light, motion blur, and dashboard/road objects(ID: 28170c86-29ba-43e8-8699-e76161f16b98);
\item \textbf{E4D-HomeWalk}: domestic walking scenes with repeated occlusions by the wearer's body or carried objects(ID: 1635447d-f96f-4f1b-8e02-faaebcd8a6d2).
\end{itemize}
We decode the video at 1~fps for testing. We reiterate that no Ego4D frame is used during pre-training: both DINO and EgoViT are trained only on our single Zurich city-walk video.

\subsection{Pseudo Ground-Truth from SAM2}
Ego4D does not provide generic bounding boxes for the main object of interest per frame. 
To obtain a rough notion of object locations without manual annotation, we use SAM2\cite{ravi2024sam} to generate a dense set of instance masks on each frame and convert them into bounding boxes.

Concretely, for each split $s \in \{\text{kitchen, driving, homewalk}\}$ and frame image $I$, we run the official Mask Generator and collect all masks whose area exceeds a small threshold to filter out tiny or noisy components.
Each mask is converted to an axis-aligned bounding box, yielding a set:
\[
\mathcal{B}_{\mathrm{SAM2}}(I) = \{b_1, \dots, b_{K}\},
\]
where $K$ is the number of boxes in frame $I$.
These boxes are stored in JSON files per split.
We found that SAM2 can under-segment cluttered regions or merge small objects, thus we treat its outputs as noisy pseudo ground-truth, primarily relying on them for visual inspection, not as a definitive quantitative benchmark.

\subsection{Zero-Shot Box Prediction and Inference Protocol}
We apply exactly the same LOST-style object discovery pipeline used in our VOC CorLoc experiments to the Ego4D-mini frames.
The tested models are:
\begin{itemize}
\item \textbf{DINO}: ViT-S/16 backbone with official DINO pre-training;
\item \textbf{EgoViT}: our proposed model (ViT-S/16).
Both pre-trained only on the single Zurich city-walk video.
\end{itemize}
The protocol involves extracting patch-level features, running the LOST algorithm to obtain predictions ($\mathcal{B}_{\mathrm{DINO}}(I)$ or $\mathcal{B}_{\mathrm{EgoViT}}(I)$), and storing the results per split.

\subsection{Qualitative Observations}

\begin{figure}[t]
  \centering
  \includegraphics[page=1,width=\linewidth]{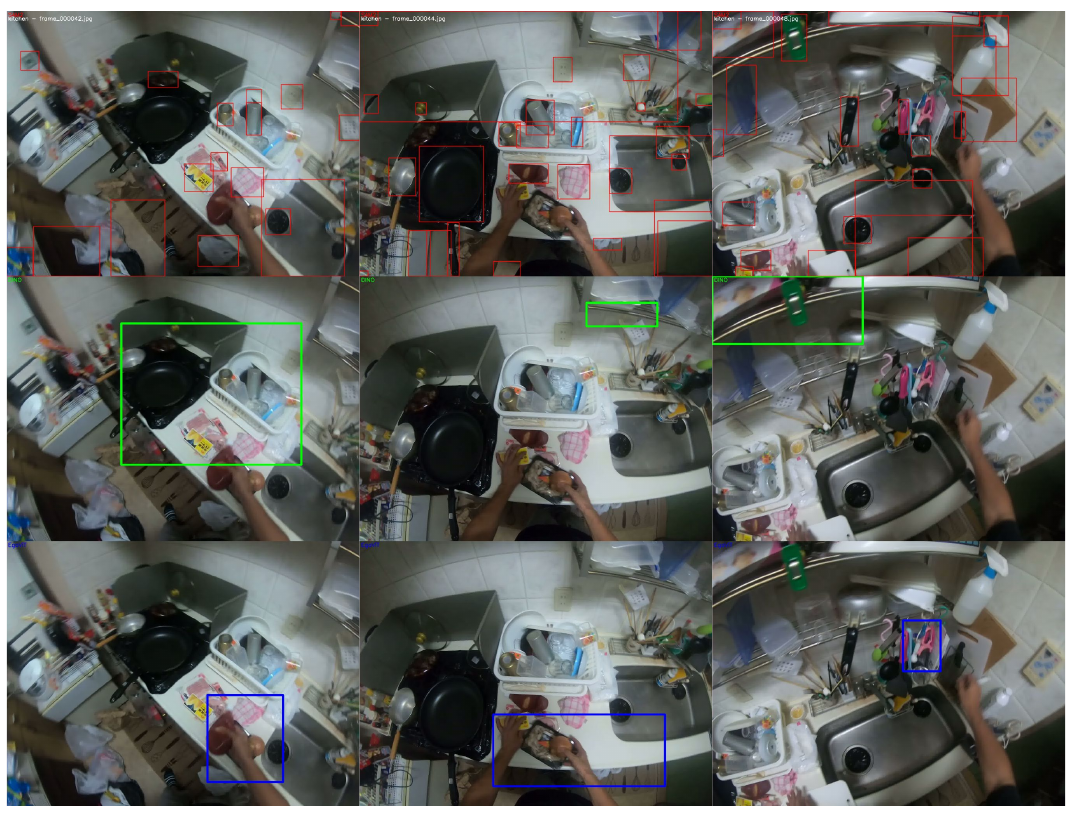}
  \caption{
  Qualitative comparison on the \textbf{Kitchen} clip.
  Each column corresponds to a different time step.
  From top to bottom: SAM2 pseudo boxes (red), DINO predictions (green), and EgoViT predictions (blue).
  EgoViT typically suppresses background clutter and localizes manipulated tools and food items near the hands, whereas DINO often locks onto larger static structures such as countertops or the sink area.
  }
  \label{fig:ego4d_kitchen}
\end{figure}

\begin{figure}[t]
  \centering
  \includegraphics[width=\linewidth]{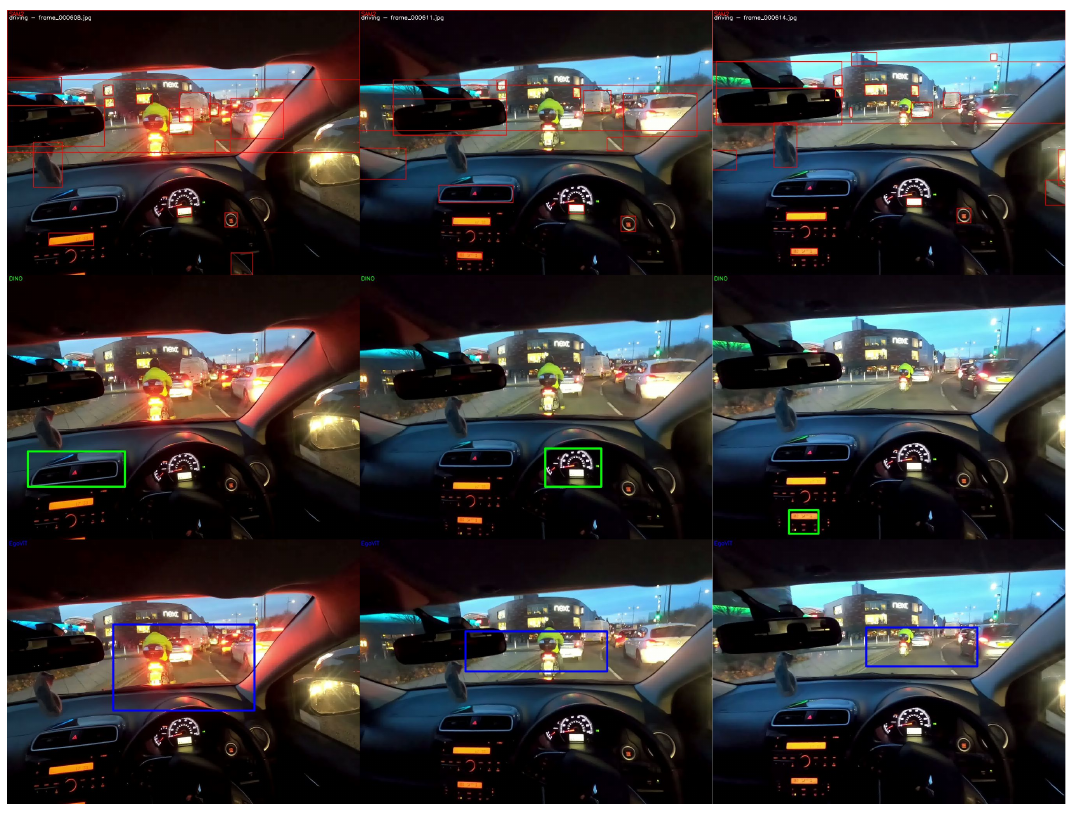}
  \caption{
  Qualitative comparison on the \textbf{Driving} clip.
  The layout is the same as in Fig.~\ref{fig:ego4d_kitchen}.
  Under low light and strong camera motion, DINO frequently drifts to dashboard edges or windshield borders, while EgoViT more reliably tracks salient traffic participants and in-cabin control elements over time.
  }
  \label{fig:ego4d_driving}
\end{figure}

\begin{figure}[t]
  \centering
  \includegraphics[page=3,width=\linewidth]{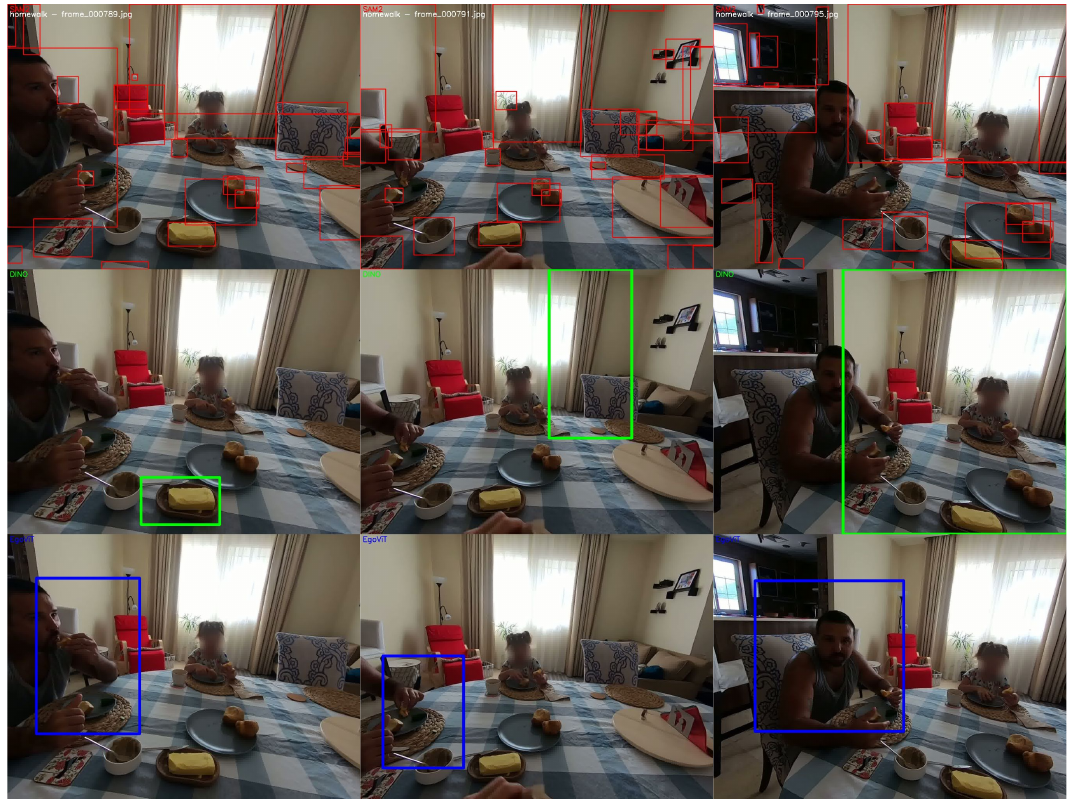}
  \caption{
  Qualitative comparison on the \textbf{HomeWalk} clip.
  The layout is the same as in Fig.~\ref{fig:ego4d_kitchen}.
  Due to frequent egocentric self-occlusions by the wearer's body and carried items, DINO often falls back to static background regions (walls, floors, furniture), whereas EgoViT maintains focus on object across occlusion and reappearance.
  }
  \label{fig:ego4d_homewalk}
\end{figure}

Each selected frame is visualized as a triplet figure: (1) SAM2 pseudo boxes (red), (2) DINO prediction (green), and (3) EgoViT prediction (blue).
We consistently observe the following patterns, which match the design goal of EgoViT:

\paragraph{More Object-Centric Localization in Cluttered Indoor Scenes.}
In the Kitchen clip (Fig.~\ref{fig:ego4d_kitchen}), DINO often locks onto large, high-contrast background structures. By contrast, EgoViT tends to place tighter boxes around manipulated tools and objects near the hands, aligning more closely with salient object regions.

\paragraph{Improved robustness in low-light, dynamic in-car scenes.}
In the Driving clip (Fig.~\ref{fig:ego4d_driving}), low-light conditions and fast camera motion make the scene challenging. We observe that DINO's predictions sometimes drift to the dashboard, windshield borders, or large textureless areas, especially when motion blur is strong. EgoViT predictions more often remain on salient foreground entities such as the leading car or steering wheel, and exhibit better temporal stability when viewed as a sequence of frames.

\paragraph{Consistent tracking of salient objects under egocentric occlusions.}
In the HomeWalk sequence (Fig.~\ref{fig:ego4d_homewalk}), the wearer's body and carried items repeatedly occlude parts of the scene. DINO occasionally switches to static background regions (walls, floors), while EgoViT more frequently maintains focus on the object being carried or manipulated across time, even when partial occlusions occur.

These trends match the design goal of EgoViT: by injecting depth cues and temporal consistency losses, the model learns to prefer depth-consistent, temporally stable proto-object regions that correspond to manipulable objects, rather than arbitrary textured patterns.

\subsection{On Quantitative CorLoc with SAM2 Pseudo Boxes}

\begin{table}[t]
\centering
\scriptsize
\caption{CorLoc (\%) on Ego4D-mini benchmark with SAM2 pseudo GT.}
\label{tab:corloc_ego4d}
\begin{tabular}{lcc}
\toprule
Split & EgoViT (\%) & DINO (\%) \\
\midrule
Kitchen   & 10.51  & 11.63  \\
Driving   & 16.61 & 12.11 \\
HomeWalk  & 25.93 & 14.89 \\
\bottomrule
\end{tabular}
\end{table}

We also experimented with a CorLoc-style quantitative evaluation where SAM2 pseudo boxes are treated as ground-truth and DINO/EgoViT boxes are counted as correct if they satisfy $\mathrm{IoU} \ge 0.5$. 
However, we found that the resulting scores are highly sensitive to SAM2's segmentation granularity.

As a result, the CorLoc numbers on these pseudo labels understate the qualitative differences visible in the visualizations and can even favor overly coarse boxes.
For this reason, we choose to present the Ego4D-mini experiment as a qualitative case study in the appendix, as shown in Table~\ref{tab:corloc_ego4d}, and reserve quantitative comparisons for datasets with human-annotated boxes and masks (PASCAL VOC, DAVIS-2017, ADE20K) in the main paper.

\paragraph{Conclusion.} 
Without any additional pre-training data, EgoViT exhibits more object-centric and temporally stable behavior than the DINO baseline when evaluated zero-shot on unseen egocentric environments. This provides strong qualitative evidence that EgoViT's geometry- and time-aware design substantially improves generalization beyond the original training video.


\end{document}